\documentclass[journal]{IEEEtran}
\ifCLASSINFOpdf
  \usepackage[pdftex]{graphicx}
\else
\fi

\usepackage{url}
\usepackage[binary-units=true]{siunitx}
\usepackage{xcolor}
\usepackage{tikz}
\usepackage[mode=buildnew]{standalone}
\usepackage{siunitx}
\usepackage{amsmath,amssymb,amsfonts}
\usepackage{nicefrac}
\usepackage{booktabs}
\usepackage{tabularx}
\usepackage{multirow}
\usepackage{enumitem}  
\usepackage{hyperref}
\usepackage{algorithm}
\usepackage{algorithmic}
\usepackage{cite}
\usepackage{verbatim}
\hypersetup{
    colorlinks=true,
    linkcolor=black,
    filecolor=magenta,      
    urlcolor=black,
}
\usepackage{color,soul}

\newcommand{\gitlink}{\href{https://github.com/eml-eda/multi-constraint-nas}{https://github.com/eml-eda/multi-constraint-nas}}

\usepackage{tikz}
\usepackage{textcomp}
\usepackage[doipre={DOI:~}]{uri}
\usepackage{lipsum}

\newcommand\copyrighttext{%
  \footnotesize \textcopyright 2021 IEEE. Personal use of this material is permitted.  Permission from IEEE must be obtained for all other uses, in any current or future media, including reprinting/republishing this material for advertising or promotional purposes, creating new collective works, for resale or redistribution to servers or lists, or reuse of any copyrighted component of this work in other works.
 
  Accepted at IEEE Transactions on Emerging Topics in Computing.
  ~\doi{10.1109/TETC.2023.3322033} }
\newcommand{\copyrightnotice}{%
\begin{tikzpicture}[remember picture,overlay,scale=0.80, every node/.style={scale=0.80}]
\node[anchor=south,yshift=10pt] at (current page.south) {\fbox{\parbox{\dimexpr\textwidth-\fboxsep-\fboxrule\relax}{\copyrighttext}}};
\end{tikzpicture}%
}

\begin{document}

\title{Enhancing Neural Architecture Search with Multiple Hardware Constraints for Deep Learning Model Deployment on Tiny IoT Devices}
%
%
%

\author{Alessio Burrello, \emph{Member, IEEE,}
        Matteo Risso, \emph{Member, IEEE,}
        Beatrice Alessandra Motetti,
        Enrico Macii, \emph{Fellow, IEEE,}
        Luca Benini, \emph{Fellow, IEEE,}
        Daniele Jahier Pagliari, \emph{Member, IEEE}
\thanks{This work has received funding from the Key Digital Technologies Joint Undertaking (KDT-JU) under grant agreement No 101095947. The JU receives support from the European Union’s Horizon Europe research and innovation programme.}%
\thanks{This work has been also supported in part by the Italian Ministry of Enterprises and Made in Italy projects D.M. MISE 31/12/2021 – Accordi per l’Innovazione – F/310164/01/X56 and F/310032/01-02/X56.}
\thanks{This work has been also supported in part by the ``TinyTrainer'' project that receives funding from the Swiss National Science Foundation under grant number No 207913.}
\thanks{We acknowledge the CINECA award under the ISCRA initiative, for the availability of high performance computing resources and support.}
\thanks{A. Burrello, M. Risso, B. A. Motetti, E. Macii, D. Jahier Pagliari are with the Politecnico di Torino, Corso Duca degli Abruzzi 24, 10129 Torino, Italy (e-mail: name.surname@polito.it). 
A. Burrello and L. Benini are with the Department of Electrical, Electronic and Information Engineering (DEI) of University of Bologna, Viale del Risorgimento 2, 40136 Bologna, Italy (e-mail: name.surname@unibo.it).
L. Benini is with the Integrated Systems Laboratory (IIS) of ETH Z\"urich, ETZ, Gloriastrasse 35, 8092 Z\"urich, Switzerland (e-mail: name.surname@iis.ee.ethz.ch).}%
\thanks{Copyright (c) 2021 IEEE. Personal use of this material is permitted. However, permission to use this material for any other purposes must be obtained from the IEEE by sending a request to pubs-permissions@ieee.org.}%
\thanks{Manuscript received xxx; revised xxxx.}}

\markboth{IEEE Transactions on Emerging Topics in Computing,~Vol.~xx, No.~yy, xxx}%
{A. Burrello \MakeLowercase{\textit{et al.}}: {Enhancing Neural Architecture Search with Latency and Memory Constrains for Tiny Deep Learning Deployment on Edge Devices}}
%



\maketitle

\copyrightnotice

\begin{abstract}
The rapid proliferation of computing domains relying on Internet of Things (IoT) devices has created a pressing need for efficient and accurate deep-learning (DL) models that can run on low-power devices. However, traditional DL models tend to be too complex and computationally intensive for typical IoT end-nodes. To address this challenge, Neural Architecture Search (NAS) has emerged as a popular design automation technique for co-optimizing the accuracy and complexity of deep neural networks. Nevertheless, existing NAS techniques require many iterations to produce a network that adheres to specific hardware constraints, such as the maximum memory available on the hardware or the maximum latency allowed by the target application.

In this work, we propose a novel approach to incorporate multiple constraints into so-called Differentiable NAS optimization methods, which allows the generation, in a single shot, of a model that respects user-defined constraints on both memory and latency in a time comparable to a single standard training. The proposed approach is evaluated on five IoT-relevant benchmarks, including the MLPerf Tiny suite and Tiny ImageNet, demonstrating that, with a single search, it is possible to reduce memory and latency by 87.4\% and 54.2\%, respectively (as defined by our targets), while ensuring non-inferior accuracy on state-of-the-art hand-tuned deep neural networks for TinyML. 

\end{abstract}


\section*{Supplementary Material}
The project's code is available at: \gitlink.

\IEEEpeerreviewmaketitle

\section{Introduction}\label{sec:introduction}
Many emerging Internet of Things (IoT) computing applications are built around Deep Learning (DL) models, ranging from audio and video classification~\cite{resnet,zhang2017hello} to biosignal analysis on wearables~\cite{reiss2019deep}, natural event prediction~\cite{7913634} or to industrial assets monitoring~\cite{Cerquitelli2021}. Evidence shows that deploying Deep Neural Networks (DNNs) directly on IoT edge devices may bring higher energy efficiency, reduced network pressure, and improved data privacy with respect to a more traditional cloud-based paradigm~\cite{edge_computing_2016}. 
Performing most computations locally would in fact limit the transmissions to the cloud, through an energy-hungry, high-latency, and potentially insecure wireless channel, to just the highly compressed model output (for example, a class label), rather than a large set of possibly sensitive raw data (such as image pixels or audio samples).

However, designing effective DL solutions while respecting the tight and multifaceted constraints of IoT end nodes is a challenging endeavour. 
An optimal model should simultaneously: i) achieve high accuracy; ii) fit in the limited memory of the device (down to few MBs of Flash and RAM for Microcontrollers~\cite{stm32h7}) and iii) match the real-time latency or throughput imposed by the application. Besides these global requirements, models should also be designed in a hardware (HW) aware fashion to achieve maximum efficiency~\cite{9789220}, for example, trying to ensure that the tensors processed by each layer fit entirely in the closest levels of the memory hierarchy, thus avoiding costly data transfers~\cite{dory}.
The solution space is huge and includes all possible variations of number, type, and hyper-parameters settings for the DNN layers~\cite{1000nas}. Thus, manual explorations are inevitably tainted by rules of thumb and traditions, and yield sub-optimal results.

Neural Architecture Search (NAS) tackles this problem by automating the search process, enabling a much more extensive exploration of the solution space than it would be possible by hand~\cite{1000nas,songhan_survey,hw_nas_survey}. This emerging approach led to the discovery of novel efficient DNNs in several domains~\cite{1000nas,songhan_survey,hw_nas_survey, mobilenetv3,tan2019efficientnet,song2021ppg}. Unfortunately, NAS algorithms that can accomodate arbitrary combinations of constraints leverage complex black-box optimization methods, such as Reinforcement Learning (RL) and Evolutionary Algorithms (EA), which are extremely time consuming (1000s of GPU hours), as shown for example in~\cite{mnasnet_2019}. Therefore, they are not affordable for most IoT application designers.
On the other hand, the so-called One-Shot or Differentiable NAS (DNAS) methods use gradient-based optimization to concurrently train a DNN and optimize its architecture, thus being much more lightweight. However, most DNAS minimize a loss function that accounts either for sheer task accuracy ($\mathcal{L}$), or for a linear combination of accuracy and a cost metric ($\mathcal{L} + \lambda \mathcal{R}$), where $\mathcal{R}$ may express the total number of parameters of the model, or the number of operations (OPs) per prediction~\cite{pit_tcomp, wan2020fbnetv2, cai2018proxylessnas}, etc.
With this formulation, finding a model with, for instance, a given number of parameters, involves an iterative tuning of the coefficient $\lambda$, to find the best balance between the two loss components.
Therefore, despite being referred to ``one-shot", current DNAS methods cannot generate a DNN that meets multiple hardware-related constraints in a single-shot. 

Although recent methods have tried to overcome this issue~\cite{luo2022lightnas,nayman_hardcore-nas_2021,fedorov2022udc}, going in the direction of a real one-shot search, they still considered a single cost metric. To the best of our knowledge, DNAS has not yet been employed in real-world, multi-constraint settings~\cite{1000nas, hw_nas_survey}. 

\begin{figure}[t]
  \centering
  \includegraphics[width=\columnwidth]{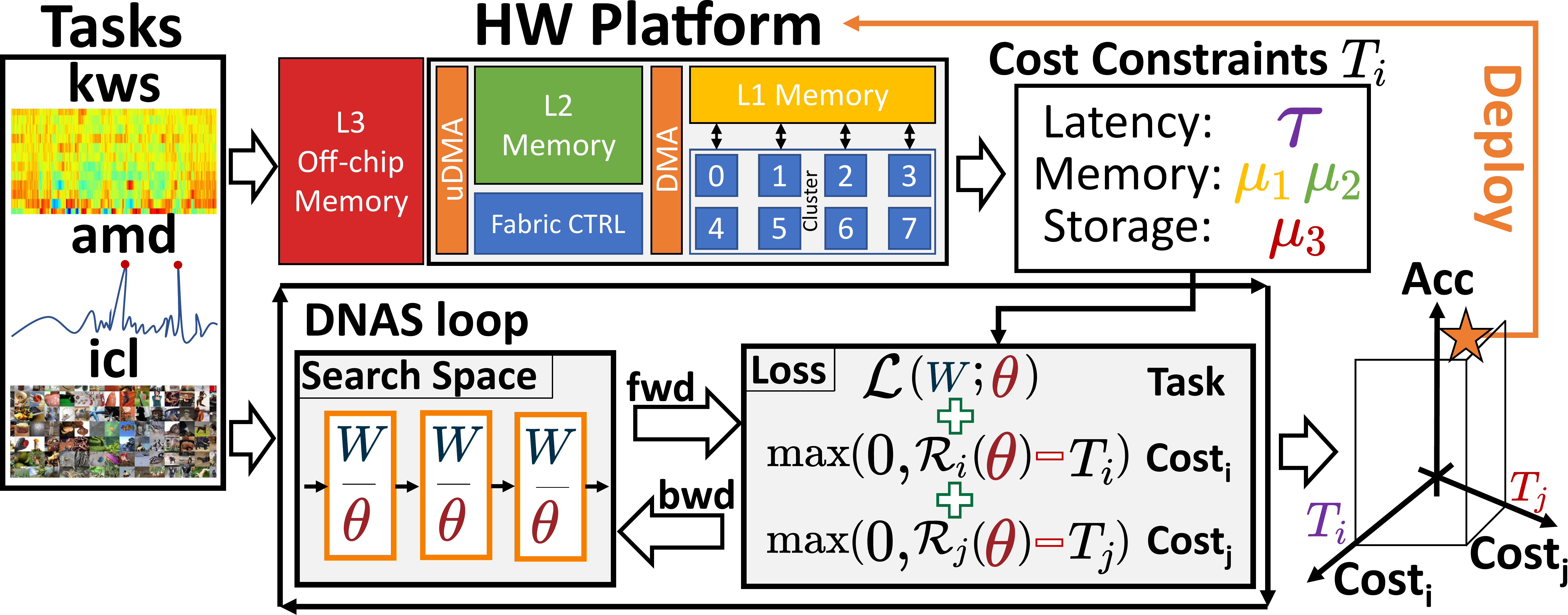}
  \vspace{-0.4cm}
  \caption{Overview of the proposed approach.}
  \label{fig:flow}
\end{figure}

In this work, which largely and highly extends~\cite{multiloss_conf}, we bridge this gap, by proposing a set of objective formulations and training recipes that collectively enable any DNAS, regardless of its specific optimization method, to produce DNNs respecting a complex set of constraints, in a time-span comparable with a single training. We name our method \textbf{D}NAS \textbf{U}nder \textbf{C}ombined \textbf{C}onstraints \textbf{I}n \textbf{O}ne-shot (\textbf{DUCCIO}). An overview of our proposed method is shown in Fig.~\ref{fig:flow}.

Specifically, the following are our main novel contributions:
\begin{itemize}
    \item We propose DNAS formulations that permit finding, in a single training run, a DNN that achieves high accuracy and respects both a total model size constraint and a maximum latency constraint (relevant for real-time tasks).
    \item We then extend this approach to finer-grain (layer-level) constraints, through which the DNN architecture can be tailored to optimally exploit the memory hierarchy of the target hardware.
    \item We identify key components of the search/training recipes that make DUCCIO effective, 
    such as progressively increasing regularizer strengths and discretized sampling. Furthermore, we show that DUCCIO can be applied to both path-based~\cite{liu2018darts} and mask-based~\cite{pit_tcomp} DNAS.
    \item We apply DUCCIO to five IoT-relevant tasks, including the entire MLPerf Tiny benchmark suite~\cite{mlperf-tiny} and Tiny ImageNet~\cite{tiny-imagenet}, using the corresponding reference DNNs as starting points. As target hardware, we focus on the ultra-low-power RISC-V-based GAP8 System-on-Chip (SoC)~\footnote{\url{https://greenwaves-technologies.com/ai_processor_GAP8/}}.
    \item Our results show that DUCCIO finds networks that reach the same or superior accuracy compared to the seed while being 55.9\% and 87.4\% smaller on 2 out of 4 tasks. On the keyword spotting task, DUCCIO reduces the size of the seed by 54.78\% while incurring only 0.43\% accuracy loss, while on anomaly detection, it reduces the size by 27.59\% with a small reduction in AUC of 1.29\%.
    We then additionally constrain the DNN latency to be 75-25\% that of the initial DUCCIO outputs, obtaining an additional set of Pareto optimal solutions.
    \item On our more complex benchmark, Tiny ImageNet, we show that DUCCIO can refine an already optimized model to match the HW memory hierarchy at layer-wise granularity, achieving an additional 54\%-61\% speedup in exchange for a 0.35\%-2.82\% accuracy drop.
    \item Lastly, we perform a set of ablation studies to demonstrate the effectiveness of our training recipes and to show that our one-shot method performs on-par with standard DNAS formulations while requiring approximately 10x less search time.
\end{itemize}

The rest of the paper is organized as follows. Section~\ref{sec:background} provides the required background, while Section~\ref{sec:related_works} surveys related works on NAS for efficient DNNs. Section~\ref{sec:methodology} describes the DUCCIO methodology in detail, and introduces the two DNAS methods on top of which we apply it in this work. Lastly, Section~\ref{sec:results} presents experimental results, and Section~\ref{sec:conclusion} concludes the paper.

\section{Background}\label{sec:background}
\subsection{Neural Architecture Search (NAS)}

In the last few years, NAS has appeared as an emerging technique to automate the design of DL models,  with a strong push from both academia and industry~\cite{1000nas,songhan_survey,hw_nas_survey}. The final objective of this research will be to integrate advanced NAS techniques in commercial AutoML platforms, which are currently still based on simpler exhaustive or random hyper-parameter tuning algorithms~\cite{edge_impulse,qeexo}. Some recently published surveys~\cite{1000nas,songhan_survey,hw_nas_survey} have tried to summarize the quickly changing landscape of existing NAS solutions, organizing them according to three dimensions: i) the search space, ii) the search strategy, and iii) the evaluation strategy.

The search space defines the possible architectural alternatives that a NAS can explore.
Literature works have considered various search space definition strategies, including: coarse spaces with global hyper-parameters decisions (number of layers, global width-multipliers, etc.)~\cite{tan2019efficientnet, mobilenetv3}; cell-based spaces with non-fixed number and type of cells~\cite{nas_reinforcement_2016, liu2018darts} where a cell is a specific combination of one or more layers; fine-grain layer-wise tuning of hyper-parameters such as the number of channels, filter size, dilation, etc.~\cite{pit_tcomp, gordon2018morphnet}.
The number of solutions in the search space can be as high as $10^{32}$~\cite{pit_tcomp},
and each solution is univocally determined by an encoding, which is method-specific, and tightly coupled with the search strategy.

The search strategy is the optimization engine used by the NAS to select good models from the search space.
An obvious measure of goodness is predictive performance (for example, classification accuracy). However, modern hardware-aware NAS also consider non-functional metrics,
such as
memory occupation, number of operations per inference, latency, etc.~\cite{hw_nas_survey}.

The way in which each DNN is scored during a search defines the evaluation strategy. For predictive performance, the obvious approach is a full training and testing of the model on a reference dataset. Similarly, non-functional metrics can be measured most accurately by means of a complete deployment of each DNN on the hardware~\cite{mnasnet_2019}. However, both these strategies are extremely time-consuming, which led to the development of so-called proxies, detailed below.

Two main families of NAS search strategies can be identified in literature, referred to as black-box and one-shot, respectively.
Early NAS approaches were mainly based on black-box engines, such as Reinforcement Learning (RL)~\cite{nas_reinforcement_2016} and Evolutionary Algorithms (EA)~\cite{nas_evol}.
In these methods, one or more architectures are iteratively sampled from the search space and evaluated in terms of all (functional and non-functional) metrics of interest.
Based on the result of such evaluation, the sampling policy is updated, and the entire process is repeated.
Black-box solutions are very effective and flexible, as they can accommodate any set of constraints and optimization targets.
However, due to the repeated sequence of sampling, training, and evaluation, they do not scale well with the dimension of the search space, requiring thousands of GPU hours for a single search~\cite{mnasnet_2019}.

The search complexity of black-box methods can be reduced thanks to ``proxy" evaluation strategies. In particular, predictive performance can be extrapolated from learning curves, testing on a single or few minibatches, or epochs~\cite{1000nas,elskenNeural2019}. Similarly, full deployment can be replaced by analytical or simulative cost models for non-functional metrics~\cite{hw_nas_survey}. However, accuracy extrapolation has been demonstrated to be unreliable, and even with proxies, black-box techniques remain extremely computationally demanding, which limits their usability by most IoT system designers~\cite{cai2018proxylessnas}.

To alleviate this problem, pursuing the goal of democratizing DNN optimization,  a set of lightweight techniques, referred to as Differentiable NAS (DNAS) or One-shot NAS has been proposed~\cite{liu2018darts}. The latter are the main focus of this work and are discussed in more detail in the next Sec.~\ref{sec:dnas_background}.

\subsection{Differentiable Neural Architecture Search (DNAS)} \label{sec:dnas_background}
DNAS methods use gradient descent as search strategy, optimizing both the weights and the network architecture in the same training loop (hence the name one-shot).
For these algorithms, search strategy and evaluation strategy are therefore strongly linked.

Early DNASes were built upon the concept of supernets, DNNs with multiple alternative paths, each corresponding to one candidate in the search space~\cite{liu2018darts}.
For example, the supernet can be constructed replacing each layer of a reference DNN with a module that includes multiple alternatives (such as convolutions with different filter sizes).
Alternative layers are combined using a set of 
architectural weights, which are then optimized during training. The optimization assigns higher weights to the alternatives that perform better according to the metric(s) of interest.
At the end of training, a discretization stage chooses a single path, namely the one formed joining the alternatives associated with the largest architectural weights in each supernet module.
We refer to these methods as path-based DNAS.

To find accurate yet efficient architectures, DNAS tools augment the standard training loss with an additional differentiable regularization term, that encodes the network cost.
Common cost measures are the number of parameters, the number of Operations (OPs) per inference, or differentiable estimates of latency or energy consumption~\cite{cai2018proxylessnas}.
In this case, the minimization objective of a DNAS training loop becomes:
\begin{equation} \label{eq:dnas}
\min_{W, \theta} \mathcal{L}(W; \theta) + \lambda \mathcal{R}(\theta)
\end{equation}
where $W$ is the set of trainable weights of the network, $\theta$ is the set of NAS architectural weights encoding the different paths in the supernet and $\lambda$ is a scalar regularization strength that controls the relative importance between the task-specific loss and complexity loss.

Although path-based DNAS can discover optimized architectures significantly faster than black-box approaches, exploring a vast search space requires a very large supernet, which takes a lot more time and memory to train compared to a ``regular" DNN.

A better alternative in this sense is represented by so-called mask-based DNAS~\cite{wan2020fbnetv2, gordon2018morphnet, pit_tcomp}.
In these algorithms, the large multi-path supernet is replaced by a standard, single-path neural network, which is referred to as the seed network. The search space comprises all DNNs that can be obtained by varying the seed’s hyper-parameters in order to reduce its complexity (using a smaller kernel-size, a lower number of features, etc.). 
Thus, the optimization goal becomes to remove the unnecessary components from each layer, while preserving the task performance of the DNN.

The exploration of the different sub-architectures of the seed network is conducted by coupling the weights and activations of the DNN with binary architectural masks. More specifically, each mask element is associated with a  specific subset of the weights or activations (for example a channel or a slice of the filter). 
During the forward pass, the binary masks are combined with their associated weights/activations through a Hadamard product. In this way, setting a given mask value to zero effectively simulates the removal of the corresponding part of the layer from the network.
The continuous relaxation of the binary mask values is then optimized akin to the architectural weights in path-based DNAS, with the aim of reducing the network complexity, by removing unnecessary components from each layer, while maintaining accuracy. In practice, mask-based DNAS algorithms can learn which weights or activations are the most important ones, and assign higher values to the associated architectural masks to retain them in the layer. On the contrary, they can assign lower mask values to the subsets of weights and activations whose contribution is less important, and can be pruned away.
At the end of the search, the final architecture is obtained by removing all the portions of the weights or activations that are associated to zero-elements in the binarized masks.

In comparison to a regular training of the seed, the masks add very little memory and search-time overheads, thus further reducing search costs compared to a path-based DNAS.
As only reduced versions of the seed can be explored, one clear disadvantage of mask-based approaches is that the search space definition becomes less flexible.
Indeed, it is not possible, with a mask-based approach, to explore DNNs that differ from the initial architecture by depth or by the type of layers.
However, this is exchanged for a finer-grained search granularity. For instance, starting from a seed convolution with 32 output channels, it is possible to explore all reduced variants (from 1 to 32 channels) with a step of 1, 
which would be almost unattainable for replication with many paths in a supernet~\cite{pit_tcomp}.

\begin{table*}[t]
\renewcommand*{\arraystretch}{1.5}
\centering
\caption{Summary of state-of-the-art cost-aware NAS methods.}
\begin{adjustbox}{width=1\textwidth}
\begin{tabular}{@{}lllll@{}}
\toprule
  \textbf{Method} &
  \textbf{Cost} &
  \textbf{Search algorithm} &
  \textbf{Loss / Reward function} &
  \textbf{Target constraint} \\ \midrule
Lemonade~\cite{elsken2019efficient} &
  Latency (Deploy), Size, OPs &
  EA &
  - &
  Yes \\
NSGA-Net~\cite{lu2019nsganet} &
OPs &
  EA &
  - &
  Yes \\
$\mu$NAS~\cite{liberis_nas_2021} &
  Size, MACs, Peak memory &
  EA &
  - &
  Yes \\
CNAS~\cite{gambella_cnas_2022} &
  Size, MACs, Memory &
  Genetic algorithm &
  $\mathcal{Q}_{\text{task}}(W,\theta) + \sum_j \lambda_{j,1} \left( R_j(\theta) + \lambda_{j,2} \max \left( 0, \left(R_j(\theta) - T_j \right)\right)\right)$ &
  Yes \\
MNASNet~\cite{mnasnet_2019}&
  Latency (Deploy) &
  RL &
  $ \mathcal{Q}_{\text{task}}(W,\theta) \left( \frac{\mathcal{R}(\theta)}{T} \right)^{\lambda}$ &
  Yes \\
TuNAS~\cite{bender2020weight}&
  Latency (LUT) &
  RL &
  $\mathcal{Q}_{\text{task}}(W,\theta) + \lambda  \left| \frac{\mathcal{R}(\theta)}{T} - 1\right|$ &
  Yes \\
AutoTinyML~\cite{perego_autotinyml_2022} &
  Size, Memory &
  BO &
  - &
  Yes \\\hline
ProxylessNAS~\cite{cai2018proxylessnas} &
  Latency (LUT) &
  Gradient-based &
  $\mathcal{L}_{\text{task}}(W, \theta) + \lambda_1 ||W||_2^2 + \lambda_2 \mathcal{R}(\theta)$ &
  No \\
MorphNet~\cite{gordon2018morphnet} &
  Size, OPs &
  Gradient-based &
  $\mathcal{L}_{\text{task}}(W, \theta) + \lambda \mathcal{R}(\theta)$ &
  No \\
PIT~\cite{pit_tcomp} &
  Size, OPs &
  Gradient-based &
  $\mathcal{L}_{\text{task}}(W, \theta) + \lambda \mathcal{R}(\theta)$ &
  No \\
FBNetV2~\cite{wan2020fbnetv2}&
  Size, OPs &
  Gradient-based &
  $\mathcal{L}_{\text{task}}(W, \theta) + \lambda \mathcal{R}(\theta)$ &
  No \\\hline
LightNAS~\cite{luo2022lightnas} &
  Latency (MLP) &
  Gradient-based + EA &
  $\mathcal{L}_{\text{task}}(W, \theta) + \lambda \left( \frac{\mathcal{R}(\theta)}{T}-1 \right)$ &
  Yes \\
HardCoRe-NAS~\cite{nayman_hardcore-nas_2021} &
  Latency (LUT) &
  Gradient-based + ILP  &
  $\mathcal{L}_{\text{task}}(W, \theta) \text{ s.t. } \mathcal{R}(\theta) < T$ &
  Yes  \\
UDC~\cite{fedorov2022udc} &
  Size &
  Gradient-based &
  $\mathcal{L}_{\text{task}}(W, \theta) + \lambda \left| \mathcal{R}(\theta) - T \right|$ &
  Yes \\
Multi-complexity loss~\cite{multiloss_conf} &
  Size, OPs &
  Gradient-based &
  $\mathcal{L}_{\text{task}}(W, \theta) + \lambda_1 \left|\mathcal{R}_1(\theta) - T \right| +  \lambda_2\mathcal{R}_2(\theta)$ &
  Yes \\\hline
\textbf{DUCCIO (Ours)} &
  Multiple &
  Gradient-based &
  $\mathcal{L}_{\text{task}}(W, \theta) + \sum_j \lambda_j \max(0, \mathcal{R}_j(\theta) - T_j)$ &
  Yes \\
 \bottomrule
\end{tabular}
\label{tab:related_works_nas}
\end{adjustbox}
\end{table*}

\section{Related Works}\label{sec:related_works}

Some NAS techniques have as only objective the optimization of the model’s performance, measured as validation accuracy or error~\cite{nas_reinforcement_2016,liu2018darts}. Such methods are not suited for resource-constrained IoT end-nodes, for which hardware-related cost metrics should be taken into account during the optimization. As mentioned in Sec.~\ref{sec:introduction}, most often, cost metrics represent constraints to be respected, rather than objectives. In other words, the NAS should ensure that the cost metric value ($\mathcal{R}$) is lower than a target ($T$). A summary of the most relevant cost-aware NAS methods is reported in Table~\ref{tab:related_works_nas}.

For black-box NAS, accounting for hardware constraints is relatively easy due to the flexibility of the search strategy.
Some of the first approaches~\cite{elsken2019efficient, lu2019nsganet, liberis_nas_2021,gambella_cnas_2022} tackle this challenge by exploiting EA or Genetic Algorithms, evolving the population of architectures in order to form an optimal Pareto front in the performance-complexity space.
In particular, CNAS~\cite{gambella_cnas_2022} models the complexity objective as a sum of terms, one for each cost metric, computed as the maximum between 0 and the excess cost with respect to a target, multiplied by a hyper-parameter that controls the penalty magnitude.
Among RL methods, MNASNet~\cite{mnasnet_2019} tries to enforce a constraint by multiplying the task quality reward term $Q_{\text{task}}$ (e.g., accuracy) by the ratio between the network's latency, measured from a complete deployment, and the corresponding target. A hyper-parameter controls the exponent of such ratio, and can assume different values whether or not the constraint is satisfied. Therefore, cost becomes a penalty term ($\lambda \neq 0$) when $\mathcal{R} > T$, and vice versa, it disappears ($\lambda = 0$) when $\mathcal{R} \leq T$.
TuNAS~\cite{bender2020weight} defines instead the cost term as the absolute difference of the ratio between the cost and the target from the ideal value of 1. Latency is estimated by means of a LUT-based approach, and the cost component is added to the reward after being scaled by a strength factor $\lambda$, always non-positive. Importantly, with this formulation, a model that breaks the constraint is penalized as much as a model that is under the target by the same cost amount. This is based on the assumption that a model with bigger cost is usually associated to a higher accuracy \cite{bender2020weight}.
Lastly, AutoTinyML~\cite{perego_autotinyml_2022}, relies on Bayesian Optimization (BO) to define the subspace of solutions that satisfy the given constraints, and to optimize both the architecture and its parameters.

Among DNAS methods, the usual approach to integrate cost metrics into the search objective uses the formulation of Eq.~\ref{eq:dnas}. In practice, the constrained optimization problem is transformed into a single-objective optimization by means of scalarization, that is, by adding all cost terms, each weighted by a different hyper-parameter, as regularizers to the single objective function. In this way, architectures which require a high amount of memory or computation resources are penalized, and this steers the search direction towards regions of the search space with better cost vs accuracy trade-offs.

Methods that follow this scheme mainly differ in the specific cost metric expression, which in order to be optimized with gradient descent, must be a differentiable function. MorphNet~\cite{gordon2018morphnet}, PIT~\cite{pit_tcomp} and FBNetV2~\cite{wan2020fbnetv2} are three mask-based DNAS that estimate memory and time/energy cost respectively as the network's size (number of parameters), and its number of Multiply-and-Accumulate (MAC) operations per prediction (OPs), expressed as a function of the architectural parameters. 
ProxylessNAS~\cite{cai2018proxylessnas}, instead, is a path-based approach that regularizes against a LUT-based latency model. As detailed in Sec.~\ref{sec:methodology},
the main issue of these approaches is the fact that there is no guarantee that the final architecture will satisfy the cost target $T$.
Thus, to obtain a model which can be actually deployed on the target hardware, while achieving high accuracy, different $\lambda$ values have to be tested (empirically, approximately 10 distinct runs are required on average~\cite{luo2022lightnas}).

Recently, there have been some attempts to develop DNAS approaches which can effectively enforce constraints. LightNAS~\cite{luo2022lightnas}, a DNAS+EA hybrid, defines the regularization term as the distance between a models' latency and a target, rescaled by the latter. Latency is estimated through a Multi-Layer Perceptron (MLP) model trained on measurements from the hardware, and the hyper-parameter $\lambda$ is adjusted during training with gradient ascent to ensure that the constraint is enforced. In practice, if the achieved latency is higher than the target, the value of $\lambda$ will be increased in order to prioritize the cost loss term during the next search iteration, and vice versa.  Thus, the formulation becomes similar to a Lagrangian relaxation, although solved with stochastic gradient descent, given the highly non-convex DNN optimization objective and cost constraints~\cite{lagrangian}. UDC~\cite{fedorov2022udc} adopts a similar regularization formulation to constrain the number of parameters, using a constant $\lambda$ value. HardCoRe-NAS~\cite{nayman_hardcore-nas_2021} combines DNAS with Integer Linear Programming (ILP), using Block Coordinate Stochastic Frank-Wolfe Algorithm to guide the search towards an optimal architecture while remaining in a feasibility region where a latency constraint is satisfied.
In our previous work~\cite{multiloss_conf} we considered for the first time multiple cost dimensions in a DNAS optimization, where one dimension (model size) is treated as a constraint, and another (number of OPs) as an objective.

\section{Proposed Method}\label{sec:methodology}
While undoubtedly helpful, DNAS tools based on Eq.~\ref{eq:dnas} have several key limitations.
First, the optimization space for real-world DNNs is constrained on multiple dimensions, with complex interdependencies, which cannot be captured by the single cost metric modeled by $\mathcal{R}$. Choosing only one metric, such as the number of OPs, the memory footprint, or a differentiable approximation of the energy consumption or latency
~\cite{gordon2018morphnet,cai2018proxylessnas} implicitly ignores the others, resulting in sub-optimal explorations. 

Furthermore, in Eq.~\ref{eq:dnas}, complexity is seen as a secondary objective to minimize, rather than a constraint. 
This clashes with the goal of most designers, who care about obtaining the most accurate DNN that meets certain cost conditions. In other words, they want to ensure that all relevant cost metrics are simultaneously within specificed bounds, so that the DNN can fit in memory, respect real-time constraints, meet the expected battery life, etc. At the same time, they are not interested in further reducing cost below such limits.

Obtaining this with Eq.~\ref{eq:dnas} is burdensome,
even in the simplified case of a single cost metric. In fact, that formulation does not impose any limit on the value of $\mathcal{R}$, and on the other hand, the optimizer can freely continue to improve its loss function by reducing $\mathcal{R}$, at the expense of accuracy, even after reaching the desired target. In practice, designers have to repeat the entire search multiple times, while tuning the regularization strength $\lambda$, until a model with the right cost, and high accuracy, is found. Although recent works have proposed alternative formulations to embed cost targets in a DNAS search~\cite{luo2022lightnas,fedorov2022udc}, they still consider a single cost metric, and try to match it exactly, which might be sub-optimal.

This paper introduces a novel and flexible DNAS formulation and training mechanism to solve the issues above, which we call \textbf{D}NAS \textbf{U}nder \textbf{C}ombined \textbf{C}onstraints \textbf{I}n \textbf{O}ne-Shot (\textbf{DUCCIO}). DUCCIO is an agnostic and plug-and-play methodology that can be built around any DNAS with minor, yet fundamental, changes, to easily ensure that multiple cost constraints are simultaneously respected.
We analyze different scenarios where DUCCIO can be applied proficiently to:
\begin{enumerate}
    \item Search for architectures that simultaneously respect a maximum storage target and a maximum number of OPs target, where the latter is a proxy for latency.
    \item Refine an architecture with layer-wise size constraints to fit the fastest memory of the target, trading-off significant speed-ups for minimal accuracy drops.
\end{enumerate}
We demonstrate DUCCIO's flexibility by applying it on top two different state-of-the-art DNAS methods, using radically different optimization mechanisms~\cite{liu2018darts,pit_tcomp}. Note that neither scenario 1) nor scenario 2) were approachable with our preliminary work of~\cite{multiloss_conf}, which additionally, was tested only on mask-based DNAS. In the rest of the manuscript, we focus on NAS for Convolutional Neural Networks (CNNs) although DUCCIO is also applicable to other types of DNN.

The rest of this section is organized as follows.
Sec.~\ref{sec:path_search} and Sec.~\ref{sec:mask_search} introduce the two DNASes on top of which we apply our method.
Sec.~\ref{sec:multireg_loss} presents the multi-regularization loss approach at the core of DUCCIO, while
Sec.~\ref{sec:training_proc} discusses the key elements of its training procedure.

\subsection{Path-based DNAS} \label{sec:path_search}
\begin{figure}[t]
  \centering
  \includegraphics[width=.9\columnwidth]{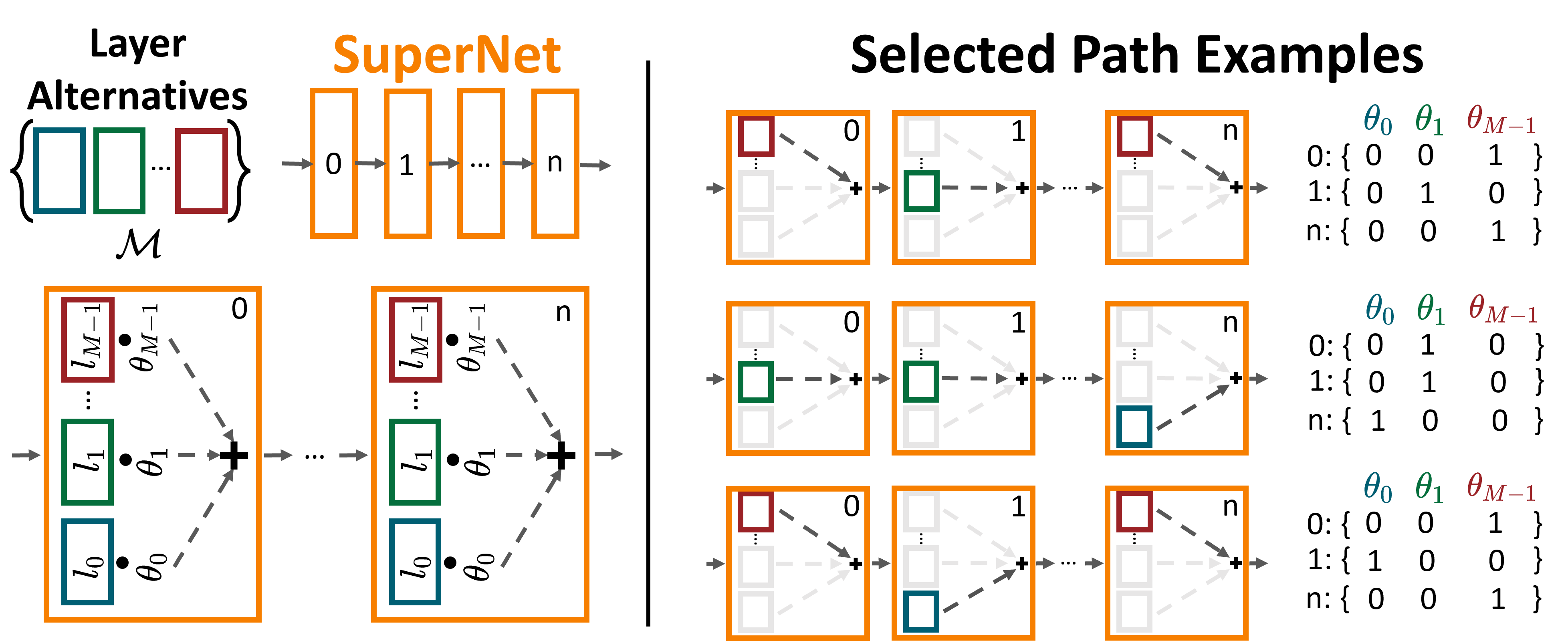}
  \caption{Target path-based DNAS method}
  \label{fig:supernet_dnas}
\end{figure}
The first DNAS that we consider is inspired by DARTS~\cite{liu2018darts}, and leverages a supernet, thus belonging to the category of path-based methods.

Fig.~\ref{fig:supernet_dnas} summarizes its optimization scheme.
The supernet is built starting from a standard CNN where each convolutional layer
$L^{(n)}$ is replaced with a module made of a set of $M$ different layer alternatives $\mathcal{M}^{(n)} = \{l_{i}^{(n)}\}_{m=0}^{M-1}$.
Within each module, all layers share the same input, while each output tensor is coupled with an additional trainable parameter $\theta_{i}^{(n)}$. 
The set of $\theta_{i}^{(n)}$ is used to select among alternatives at the end of the search.
The supernet is inserted in a normal training loop, where the weights of each layer and the corresponding $\theta$ are trained jointly, with the objective function described in Sec.~\ref{sec:multireg_loss}.

With respect to DARTS, there are two key differences in our path-based DNAS. In accordance to most recent NAS literature~\cite{snas, wan2020fbnetv2}, we leverage the Gumbel-Softmax~\cite{gumbel} sampling strategy, as opposed to the standard Softmax used in~\cite{liu2018darts}. Moreover, we use a discretized sampling strategy, where only one path is selected in each training iteration.
In practice, the output of the supernet module is built as:
\begin{equation} \label{eq:output_supernet}
y^{(n)}(x) = \sum_{i=0}^{M-1}  g(\theta^{(n)}_i) \cdot l^{(n)}_i(x)
\end{equation}
where $g(\theta^{(n)}_i)$ are discretized Gumbel-Softmax coefficients:
\begin{equation} \label{eq:gumbel_coeff}
g(\theta^{(n)}_i) = \mathrm{onehot}_i\left(\dfrac{\exp(\theta^{(n)}_i + \epsilon)}{\sum_{i} \exp(\theta^{(n)}_i + \epsilon)}\right)
\end{equation}
with $\epsilon \sim Gumbel(0, 1)$ being i.i.d. samples drawn from the Gumbel distribution~\cite{gumbel}, and onehot being a function that discretizes to 1 the maximum sampled value among the alternatives, and to 0 all the others. We use a Straight-Through Estimator (STE)~\cite{courbariaux2015binaryconnect}, replacing onehot with an identity function during backward passes. This lets gradients flow through the network despite the non-differentiable function. As detailed in Sec.~\ref{sec:training_proc}, discretized sampling is fundamental to respect cost constraints.

At the end of training, the optimized architecture is exported selecting, for each supernet module, the alternative corresponding to the largest $\theta^{(n)}_i$.

\subsection{Mask-based DNAS} \label{sec:mask_search}
\begin{figure}[t]
  \centering
  \includegraphics[width=.9\columnwidth]{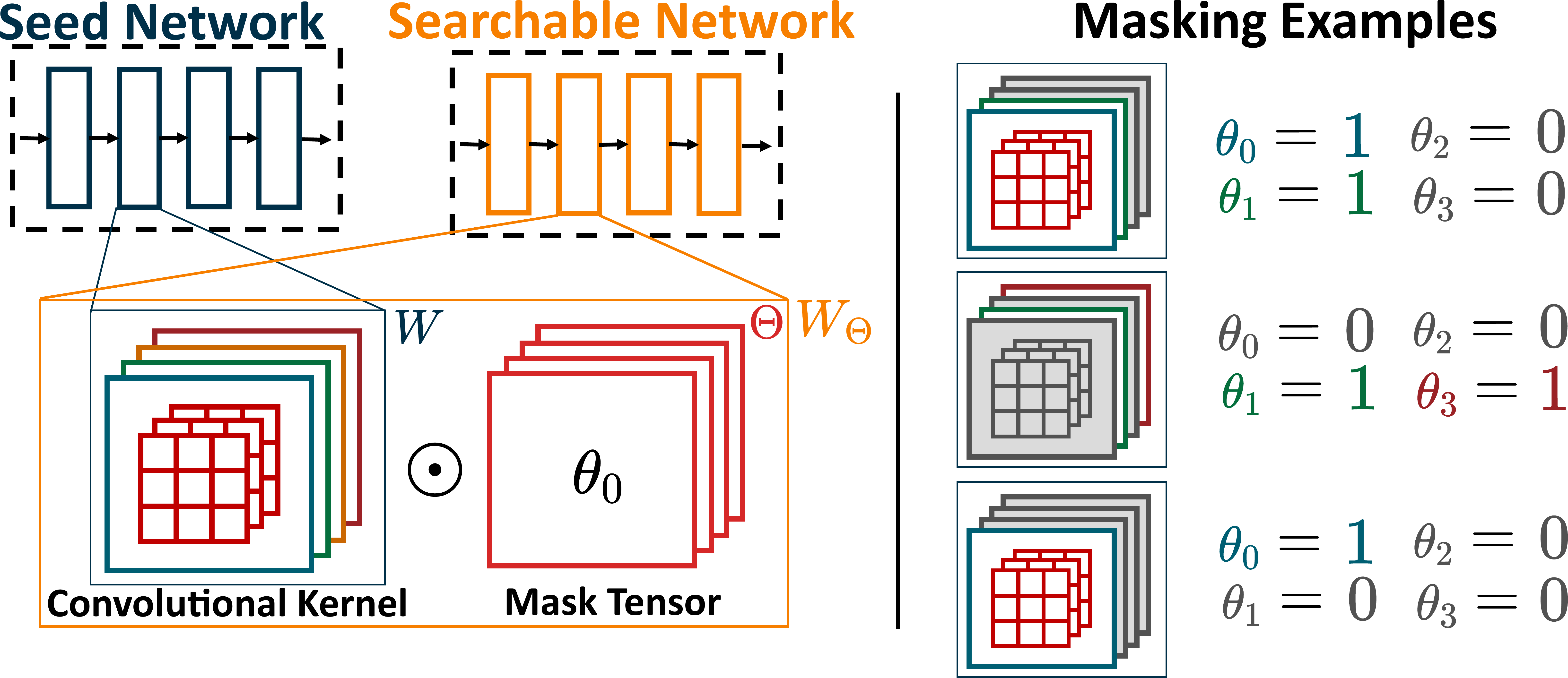}
  \caption{Target mask-based DNAS method.}
  \label{fig:search_space}
\end{figure}
The second DNAS that we consider is a simple yet powerful mask-based approach which, starting from a seed CNN, performs a fine-grained optimization of the number of output channels of all its convolutional layers.
This DNAS, schematized in Fig.~\ref{fig:search_space} is an extension to 2D CNNs of the channel-search approach originally proposed in PIT~\cite{pit_tcomp}.

In this case, a searchable model, that is, one that can be optimized by the DNAS, is built coupling each convolutional weight tensor in the seed $W^{(n)}$ with a trainable mask vector $\theta^{(n)}$.
The dimension of $W^{(n)}$ is $C^{(n)}_{out} \times K^{(n)}_x \times K^{(n)}_y \times C^{(n)}_{in}$, where $C^{(n)}_{in}$, $C^{(n)}_{out}$ are the input/output channels and $K^{(n)}_x$, $K^{(n)}_y$ are the horizontal/vertical kernel sizes. 

The corresponding
$\theta^{(n)}$ includes $C^{(n)}_{out}$ elements, one per output channel, and is used to selectively prune the least important channels during the search.
Accordingly, a masked weight tensor $W_{\Theta}^{(n)}$ is built as:
\begin{equation} \label{eq:masked_kernel}
W_{\Theta}^{(n)} = W^{(n)} \odot \mathcal{H}(\theta^{(n)})
\end{equation}
where
$\odot$ is the Hadamard product, and $\mathcal{H}$ is a Heaviside step function with a fixed threshold $t_h=0$, which has the effect of binarizing $\theta^{(n)}$ ($1 \text{ if } \theta^{(n)}_i \ge t_h, 0 \text{ otherwise}$). 
Each binarized mask element is multiplied with a slice of weights $W^{(n)}_i$ along the $C^{(n)}_{out}$ dimension. The slice contains all weights of the $i$-th convolutional filter.
Therefore, when $\mathcal{H}(\theta^{(n)}_i) = 0$, the $i$=th output channel is to all effects eliminated from the network. Conversely, the channel is kept if  $\mathcal{H}(\theta^{(n)}_i) = 1$.

The obtained searchable network is again inserted in a normal training loop, where $W$ and $\theta$ are trained together according to DUCCIO's objective function.
As before, binarization ensures that we sample a single discretized architecture in each training step (see the examples on the right of Fig.~\ref{fig:search_space}).
During backward passes, a STE based on BinaryConnect~\cite{courbariaux2015binaryconnect} replaces the Heaviside step with an identity function to make the search differentiable.

In this case, the final export of the optimized model permanently eliminates all channels that correspond to a $\theta^{(n)}_i < t_h$, shrinking each Convolutional layer accordingly.

\subsection{Multi-Constraint Loss Formulation} \label{sec:multireg_loss}
As discussed in Sec.~\ref{sec:background} and \ref{sec:related_works}, most state-of-the-art DNAS tools~\cite{cai2018proxylessnas, gordon2018morphnet, wan2020fbnetv2} optimize a variant of Eq.~\ref{eq:dnas}, which sums the task-specific loss $\mathcal{L}$ and a regularization term $\mathcal{R}$, where the latter models a single cost metric as a differentiable function of the NAS architectural parameters ($\theta$), and is scaled by a strength coefficient $\lambda$.
The specific meaning of $\mathcal{R}$ varies; some approaches~\cite{gordon2018morphnet,wan2020fbnetv2} express it as the total number of parameters or OPs in target layers (e.g., all Convolutional layers); others~\cite{cai2018proxylessnas} use more direct approximations of the end-to-end latency or energy consumption of the obtained DNNs. The limitations of these approaches have been highlighted at the beginning of this section. Namely:
\begin{enumerate}
    \item They consider a single cost metric, whereas real systems are constrained in multiple dimensions.
    \item They cannot guarantee that the cost estimate $\mathcal{R}$ is lower than a target value.
    \item Even when the target is reached, $\mathcal{R}$ remains part of the objective function, hence the optimizer continues to prefer lower-cost models, although this is not necessary.
\end{enumerate}
Therefore, sweeping $\lambda$, trying tens of different values at least, is the only way to generate models with the desired cost and accuracy characteristics.

Recently, UDC~\cite{fedorov2022udc}, LightNAS~\cite{luo2022lightnas} and our previous work of~\cite{multiloss_conf} added an explicit cost target $T$ in the regularization term, and let the optimizer reduce the difference between the estimated cost and the target, by adding a term proportional to $|\mathcal{R}(\theta) - T|$, or equivalent, to the loss function.

Although making an important step forward, these methods do not solve completely the issues above. First, except for~\cite{multiloss_conf} they still consider a single cost dimension, thus not solving issue 1). Similarly, issue 3) is also not eliminated, since once $\mathcal{R}(\theta)$ becomes $<T$, the new loss component starts to have an opposite effect, favoring more costly models to reduce the difference from the target. Some papers~\cite{bender2020weight} claim that this is not a problem, since more complex models tend to also be more accurate. However, our experiments demonstrate that this is not always the case, and that letting the NAS freely change the model's cost within the boundaries imposed by constraints yields superior results. In practice, we sometimes find lower-complexity models that perform equally well with those that closely match the target.

DUCCIO addresses the limitations of previous DNAS tools proposing the following objective function formulation:
\begin{equation} \label{eq:duccio}
\min_{W, \theta} \mathcal{L}(W; \theta)  + \sum_{j=0}^J \lambda_j \max(0, \mathcal{R}_j(\theta) - T_j)
\end{equation}
which includes two fundamental new elements. First, we consider an arbitrary number ($J$) of cost metrics ($\mathcal{R}_j$) and corresponding constraints ($T_j$), each with its own regularization strength. Second, we use a $\max$ function to ensure that the loss is penalized only when the cost metric exceeds the constraint. Vice versa, whenever $\mathcal{R}_j(\theta) < T_j$, the $j$-th cost component disappears from the loss, thus not impacting the gradients of NAS parameters. The strength coefficients $\lambda_j$ are scheduled during a DUCCIO search to ensure that, at the end of training $\mathcal{R}_j(\theta) < T_j,\ \forall j$, as discussed in Sec.~\ref{sec:training_proc}.

Importantly, nothing prevents from adding cost objectives (in the form of an additive term $+\lambda_j \mathcal{R}_j(\theta)$) to the loss function of DUCCIO. This can be done similarly to what was presented in our previous work of~\cite{multiloss_conf}.
One reason to do so could be that, for a given cost metric (say energy consumption) there is not a specific constraint to be matched, but rather, the designer wishes to explore the entire Pareto front of trade-offs with accuracy. Then, for that specific dimension, DUCCIO would become equivalent to a standard DNAS. However, in this work, we focus on true one-shot DNAS, hence we consider all cost dimensions as constraints.

Next, we describe two practical applications of DUCCIO's loss formulation:

\subsubsection{Multiple Global Constraints}
A first straight-forward use case is to impose multiple global constraints, relative to the entire model under optimization. Often, system designers must ensure that: i) the total DNN size fits the storage space available on the target and ii) the total inference latency is below a real-time constraint. In this case, Eq.~\ref{eq:duccio} can be instantiated as:
\begin{equation} \label{eq:duccio_global}
\min_{W, \theta} \mathcal{L}(W; \theta)  + \lambda_0 \max(0, \mathcal{O}(\theta) - T_o) + \lambda_1 \max(0, \mathcal{S}(\theta) - T_s)
\end{equation}
where $\mathcal{S}$ models the size (storage footprint) of the DNN as a function of the NAS parameters $\theta$, and $\mathcal{O}$ is a proxy for latency.
To estimate the total size of the model, we sum the sizes of all NAS target layers:
\begin{equation} \label{eq:size_cost}
\mathcal{S}(\theta) = \sum_{n=0}^{N-1} \mathcal{S}^{(n)}(\theta)
\end{equation}
where the expression for $\mathcal{S}^{(n)}$ depends on the chosen DNAS. For our target path-based method, we estimate it as the number of weights in each supernet alternative, multiplied with the corresponding NAS sampling value:
\begin{equation} \label{eq:size_darts}
\mathcal{S}^{(n)}(\theta) = \sum_{i=0}^{M-1} g(\theta^{(n)}_i) \cdot \mathcal{S}^{(n)}_i
\end{equation}
where for example $\mathcal{S}^{(n)}_i = C_{in}^{(n)} \cdot C_{out}^{(n)}\cdot K_{x}^{(n)} \cdot K_{y}^{(n)}$ for a standard convolution, or $\mathcal{S}^{(n)}_i = C_{in}^{(n)} \cdot K_{x}^{(n)} \cdot K_{y}^{(n)} + C_{in}^{(n)} \cdot C_{out}^{(n)}$ for a sequence of DepthWise + PointWise convolutions such as those present in MobileNets~\cite{howard2017mobilenets}.

For the mask-based DNAS, instead, we estimate model size as the number of weights in non-masked filters. Taking a standard convolution again as an example, we have:
\begin{equation} \label{eq:size_pit}
\mathcal{S}^{(n)}(\theta) = C_{in}^{(n)}(\theta) \cdot C_{out}^{(n)}(\theta) \cdot K_{x}^{(n)} \cdot K_{y}^{(n)}
\end{equation}
where the dependency of $C_{out}$ from $\theta$ comes from the fact that channels with $\theta^{(n)}_i < t_h$ are to all effects eliminated from the model. $C_{in}$ also changes indirectly as a result of the shrinking $C_{out}$ in previous layers. For instance, for a simple sequential model $C_{in}^{(n)} = C_{out}^{(n-1)}$.

For what concerns latency, a 
basic
model can simply count the number of Multiply-and-Accumulate (MAC) OPs in each target layer. For convolutions, this can be expressed as:
\begin{equation} \label{eq:ops_cost}
\mathcal{O}(\theta) = \sum_{n=0}^{N-1} \mathcal{S}^{(n)}(\theta) \cdot O^{(n)}_{x} \cdot O^{(n)}_{y}
\end{equation}
where $O^{(n)}_{x}$ and $O^{(n)}_{y}$ are, respectively, the output feature map width and height of the n-th convolutional layer.
Note that while $\mathcal{O}$ and $\mathcal{S}$ are linked, optimizing one or the other is not identical. In particular, for most DNNs, output feature sizes tend to reduce while going forward in the network, due to the effect of pooling layers, strided convolution, etc., whereas the number of channels tends to increase. Thus, initial layers often contribute more to latency, and final layers more to size.

We use this simplified model in our work, since it correlates sufficiently well with actual latency for our target platform. However, DUCCIO is orthogonal to the specific cost expression, and would work as well with more accurate differentiable latency estimates, such as those proposed in~\cite{cai2018proxylessnas,luo2022lightnas}.

\subsubsection{Layer-wise Constraints}\label{sec:duccio_layerwise}
Another common scenario for DNN systems' engineers is to have a DNN model that is not ideally sized for the available hardware target. For instance, the DNN might include one or more layers that slightly exceed one particular memory level (for instance, the L2) thus requiring costly transfers of input, output, or weights tiles from the next level (L3)~\cite{dory}. With DUCCIO, we can perform a one-shot optimization that shrinks all and only these layers, while maximizing accuracy, by formulating the objective as:
\begin{equation} \label{eq:duccio_local}
\min_{W, \theta} \mathcal{L}(W; \theta)  + \sum_{n \in Crit.} \lambda_n \max(0, \mathcal{M}^{(n)}(\theta) - T_m)
\end{equation}
where $Crit.$ is the set of critical layers, which are those slightly overflowing a given memory level, and $T_m$ is that level's available space 
(that is, its total size minus a constant offset that accounts for code size, other variables, etc.). $\mathcal{M}^{(n)}(\theta)$ is the $n$-th layer's total memory which, for a Convolution, can be expressed as:
\begin{equation} \label{eq:layer_mem_cost}
\mathcal{M}^{(n)}(\theta) = \mathcal{S}^{(n)}(\theta) + I^{(n)}_{x} I^{(n)}_{y} C^{(n)}_{in}(\theta) + O^{(n)}_{x} O^{(n)}_{y} C^{(n)}_{out}(\theta) 
\end{equation}
where the three addends account for the memory required by parameters, inputs, and outputs, respectively, and $I^{(n)}_{x}$, $I^{(n)}_{y}$ are the input feature dimensions. The rightmost two addends depend on $\theta$, through $C_{in}$ and $C_{out}$ only in case of a mask-based DNAS, while they are fixed for the path-based method.

Note that, in Eq.~\ref{eq:duccio_local}, the constraints are imposed on each layer's memory, differently from Eq.~\ref{eq:duccio_global}, where the added loss terms only limit the aggregated metrics values over the entire DNN. Furthermore, while the equation includes a common target $T_m$ for all critical layers, for notational simplicity, in general we could also support different targets, to ensure some layers fit in L1 memory, while others fit in L2, etc.

To the best of our knowledge, none of the state-of-the-art DNAS works have considered such type of fine-grain, hardware-aware constraints. Our results of Sec.~\ref{sec:results} show that tailoring each layer to the memory hierarchy in this way can lead to significant speedups, in exchange for minor accuracy drops.

\subsection{Training Procedure} \label{sec:training_proc}
Alg.~\ref{alg:nas_search} reports an overview of a DUCCIO training procedure. As it is common in DNAS, we have three separate phases of warmup, search and fine-tuning.
We start with a warmup, which consists of a normal training of the full supernet (for the path-based DNAS) or seed network (for the mask-based one). In this phase, only the normal weights $W$ are trained. NAS parameters $\theta$ are frozen at their initialization value (1.0) to ensure that all paths/channels are trained. Specifically, for the path-based DNAS, this results in a uniform sampling of all supernet paths, whereas for the mask-based method, all channels of the seed model are kept active.
In this phase, the gradients are computed only with respect to the task-dependent loss function $\mathcal{L}$.
Noteworthy, warmup results are independent from the imposed constraints, thus they can be saved and reused in case of multiple searches (for example, to target different hardware).
\renewcommand{\algorithmiccomment}[1]{$\triangleright$#1}
\begin{algorithm}[t]
\begin{algorithmic}[1]
\footnotesize
\caption{\label{alg:nas_search}}
    \FOR[warmup loop]{$e \gets 1, \dots, \rm Epochs_{wu}$}
    \STATE Update $W$ based on $\nabla_{W} \mathcal{L}(W)$
    \ENDFOR
    \STATE $e \gets 1$
    \WHILE[search loop]{ $e < \rm Epochs_{sr}$ \textbf{or} not converged}
        \STATE Update $W, \theta$ based on $\nabla_{W, \theta} (\mathcal{L}(W; \theta) + \sum_{j=0}^J \lambda_j \max(0, \mathcal{R}_j(\theta) - T_j)$
        \STATE Schedule $\lambda_j,\ \forall j$
        \STATE $e \gets e + 1$
    \ENDWHILE
    \FOR[fine-tuning loop]{$e \gets 1, \dots, \rm Epochs_{ft}$}
        \STATE Update $W$ based on $\nabla_{W} \mathcal{L}(W)$
    \ENDFOR
\end{algorithmic}
\end{algorithm}

The second phase consists of the actual architecture optimization.
In this step, the weights $W$ and the NAS parameters $\theta$ are optimized together, to minimize one embodiment of the DUCCIO objective of Eq.~\ref{eq:duccio}.
We run this loop for a minimum number of epochs ($Epochs_{sr}$) and then continue until convergence, with an early-stop mechanism that monitors the task loss $\mathcal{L}$ on an unseen validation-split, and stops the search when the loss
stops improving.
When a default validation set is not provided by a specific benchmark, we generate it by randomly sampling 10\% of the training set. 

At the end of the search phase, we export the model resulting from the final $\theta$ values, as detailed in Sec.~\ref{sec:path_search} and \ref{sec:mask_search}. We then perform a fine-tuning of this model in which, similarly to warmup, we only train the weights $W$ based on the task-specific loss $\mathcal{L}$.
In all our experiments, we keep the number of warmup and fine-tuning epochs equal to the ones reported in the MLPerf Tiny official repository (see Sec.~\ref{sec:benchmarks} for details).

In the rest of this section, we discuss two key aspects that make the procedure described in Algorithm~\ref{alg:nas_search} effective to realize a one-shot DNN optimization under multiple constraints.

\subsubsection{Discretized Sampling}
At the end of a DNAS search, regardless of the specific method, a single discrete DNN architecture is sampled from the solution space and returned. In a multi-constraint setup, we want to force that this architecture respects $R_j(\theta) < T_j,\ \forall j$. To this end, we claim that we shall not apply a continuous relaxation of the sampling problem during the search, as done in~\cite{liu2018darts} and many other popular DNAS works. We show this with a simple example, consisting of a DNN with a single NAS target layer, to which we impose a maximum storage size target of $T_j = 600$. A path-based DNAS might include two alternatives for this target layer, with $S^{(1)}_0 = 100$ and $S^{(1)}_1 = 800$, respectively. In the original formulation of~\cite{liu2018darts}, these two would be linearly combined as in Eq.~\ref{eq:size_darts}, but using continuous NAS parameters, produced by a standard SoftMax operation. In this case, at the end of a search loop, the optimization could produce $g(\theta^{(n)}_0) = 0.49$ and $g(\theta^{(n)}_1) = 0.51$, which in turn yields: $S^{(n)}(\theta) = 0.49 \cdot 100 + 0.51 \cdot 800 \approx 450 < T_j$. This solution respects the constraints (in the continuous relaxation domain), thus being optimal with respect to the cost component of the loss function. However, given that the SoftMax is monotonic, and that the final architecture is extracted from the supernet based on the largest $\theta$, the DNAS will eventually select the second alternative, with a size of 800, which clearly violates the constraint.

In DUCCIO, we solve this problem by performing a discrete sampling (i.e., selecting a single, discrete architecture) in each training iteration. For the path-based DNAS, this is realized by the onehot function in Eq.~\ref{eq:gumbel_coeff}, while in the mask-based case, it is obtained thanks to the Heaviside step in Eq.~\ref{eq:masked_kernel}. We note that this is different from both original papers, since also \cite{pit_tcomp} used the continuous relaxation for cost (size or OPs) estimation. However, other works, such as~\cite{cai2018proxylessnas,bender_droput} use discretized sampling, showing that this is also beneficial to improve the correlation between the accuracy of the model during search and after the final export.

A possible alternative method to reduce the mismatch between the cost estimated during search and the one of the final model in path-based DNAS, while maintaining the continuous relaxation, is to polarize the $\theta$ arrays, ensuring that their values are not close to each other. This can be achieved with an additional loss component proportional to the Inverse Coefficient of Variation (ICV)~\cite{shazeerOutrageously2017}: $\nicefrac{\mu(g(\theta))}{\sigma(g(\theta))}$ where $\mu$ and $\sigma$ are the mean and standard deviation respectively, and $g()$ is a standard SoftMax or Gumbel SoftMax, without discretization.
We tested this approach as well, but our results show that discretized sampling produces higher quality results.

\subsubsection{Regularization Strength Scheduling}
In our previous work of~\cite{multiloss_conf}, we showed that a size target (expressed through the absolute value difference formulation) could be enforced by imposing a large and constant regularization strength $\lambda$, thus removing the need for an iterative tuning of this coefficient. While this remains true, in this work we show that setting a large $\lambda$ at the beginning of a search might yield sub-optimal results in terms of accuracy, partly due to our new formulation based on a max() function. In fact, for some benchmarks, we empirically verified that the search converges too fast towards architectures with a cost lower than the constraint, but possibly corresponding to a sub-optimal local minimum in terms of accuracy.

To solve this issue, we implement a scheduling of the $\lambda_j$ of Eq.~\ref{eq:duccio}, identical for all considered constraints. Namely, we set the target strength as in our previous work:
\begin{equation}\label{eq:lambda_target}
    \lambda_{j,target} = \frac{\hat{\mathcal{L}}}{|\hat{\mathcal{R}}_{j} - T_j|}
\end{equation}
where $\hat{\mathcal{L}}$ and $\hat{\mathcal{R}}$ are the task loss and cost at the end of warmup. This ensures that, for cost metrics that are above their corresponding target, the values of the addends in the summation of Eq.~\ref{eq:duccio} eventually become comparable with the task-loss term. However, rather than setting this value from the beginning, we increase it linearly during the initial fixed number of epochs of the search phase. In practice, at each epoch, we have:
\begin{equation}
    \lambda_{j} = \min \left(\frac{e\cdot \lambda_{j, target}}{\rm Epochs_{sr}}, \lambda_{j, target}\right)
\end{equation}

Our results show that this scheduling yields more stable results compared to a fixed $\lambda_j$, in varying combinations of task, DNAS method, and cost targets.

\section{Results}\label{sec:results}

\begin{figure*}[t]
  \centering
  \includegraphics[width=0.9\textwidth]{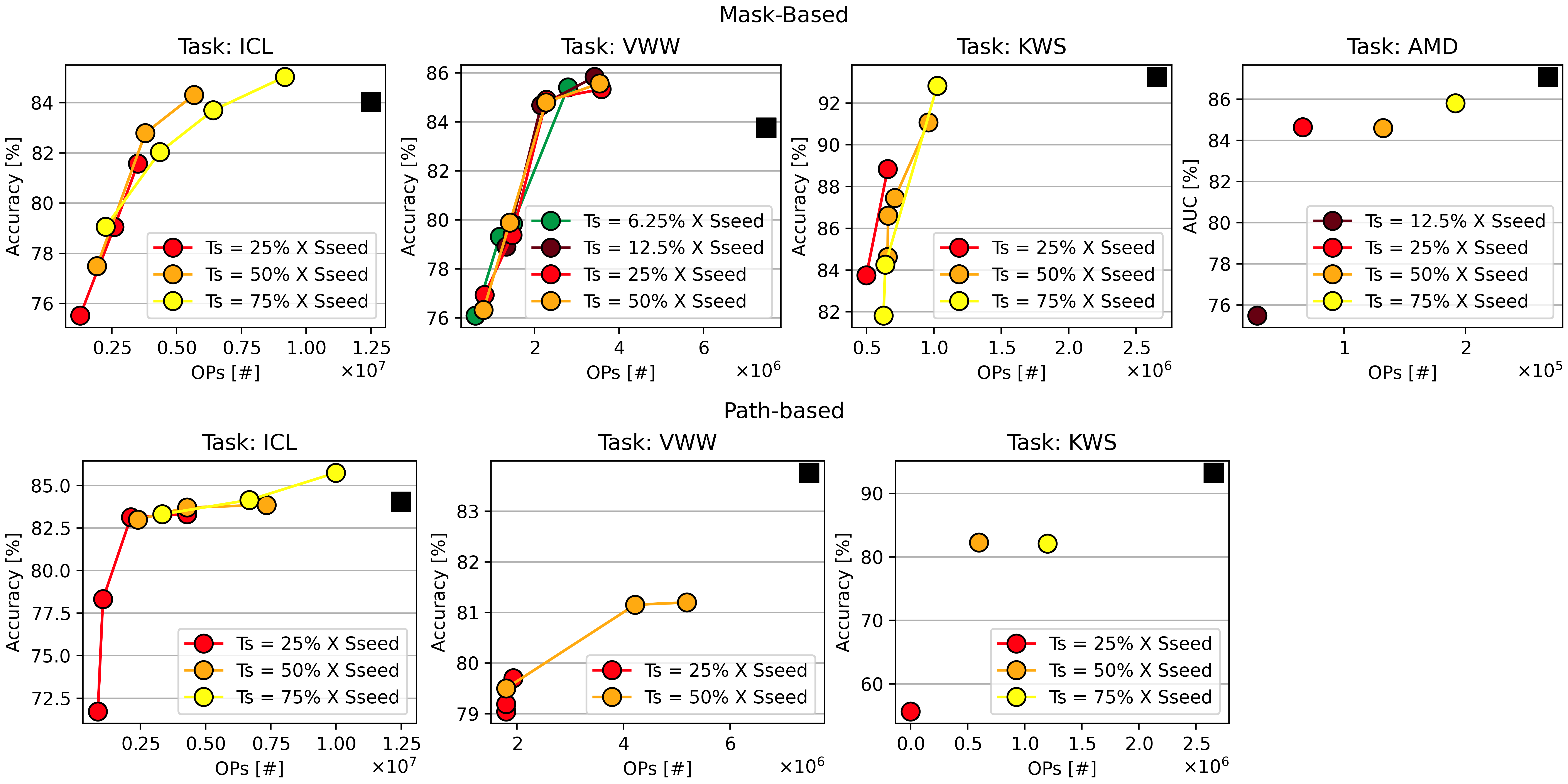}
  \caption{Accuracy versus OPs results for different size targets. In the top row, the results obtained by applying a mask-based DNAS algorithm. In the bottom one, the results obtained by using a path-based approach.}
  \label{fig:pareto_fronts}
\end{figure*}
\subsection{Experimental Setup}

We implemented DUCCIO in Python 3.10.9 and PyTorch 1.13.1, on top of the open-source PLiNIO library\footnote{\url{https://github.com/eml-eda/plinio}}. To benchmark it, we considered five edge-relevant tasks, including the four tasks of the MLPerf Tiny suite~\cite{mlperf-tiny} and image classification on the Tiny ImageNet~\cite{tiny-imagenet} dataset.
For each task, a reference DNN is identified, which is used as seed architecture for the mask-based DNAS of Sec.~\ref{sec:mask_search} and as blueprint to construct the supernet for the path-based DNAS of Sec.~\ref{sec:path_search}.
In particular, the supernet is built by substituting all convolutional layers of the reference DNN with a selection among four alternatives: i) convolution with a $3 \times 3$ filter, ii) convolution with a $5 \times 5$ filter, iii) depthwise-separable convolution, i.e., the sequence of a $3 \times 3$ depthwise convolution and a $1 \times 1$ pointwise convolution \cite{howard2017mobilenets}, and iv) identity operation. The fourth option is only included for layers that have stride $=1$ and $C_{in} = C_{out}$.

On the MLPerf Tiny suite tasks, we tested DUCCIO with global memory and OPs constraints, using both mask-based and path-based DNAS on each benchmark, with the exception of the Anomaly Detection one, since the reference architecture is a fully-connected (FC) autoencoder, which does not support the choice of different layer alternatives.
Furthermore, we used Tiny ImageNet to test the layer-wise memory-aware refinement procedure discussed in Sec.~\ref{sec:duccio_layerwise} with the mask-based DNAS, since it permits finer-grain memory optimizations.

\subsubsection{Benchmarks} \label{sec:benchmarks}
The Image Classification (ICL) task targets the CIFAR-10 dataset\cite{krizhevsky2009learning}, which includes 60,000 32x32x3 RGB images divided into 10 classes. We used the full CIFAR-10 test set in our experiments, rather than the reduced set proposed by the MLPerf Tiny suite, to have more stable results.
The seed DNN is a ResNet-like~\cite{resnet} architecture with a backbone of 8 convolutional layers. The number of training epochs (which is also used for warmup and fine-tuning during NAS) is 500.
Visual Wake Words (VWW) is a binary classification task where the goal is to determine if there is at least one person in the input image. It is based on the MSCOCO 2014 dataset\cite{chowdhery2019visual}, which contains more than 100k 96x96x3 RGB images.
The reference CNN is a MobileNetV1~\cite{howard2017mobilenets} with a width multiplier of $0.25$, which is trained for 20 epochs.
The KeyWord Spotting (KWS) benchmark considers the Speech Commands v2 dataset\cite{warden2018speech}, with 105,829 utterances, to be classified into 12 classes, 10 words and two special labels (``unknown" and ``silence").
The reference architecture is the Depthwise Separable CNN (DS-CNN) presented in~\cite{zhang2017hello}. The number of training epochs is set to 36.
The last MLPerf Tiny benchmark is Anomaly Detection (AMD), which targets the Toy-car data fold included in the DCASE2020 dataset\cite{koizumi2020description}, to detect anomalies based on machine operating sounds. The reference model is a simple FC Autoencoder. In this case, the number of training epochs is 100.
Lastly, we consider a more challenging image classification task based on Tiny ImageNet~\cite{tiny-imagenet}, with 100k 64x64x3 RGB images to be classified into 200 classes with a ResNet18~\cite{resnet} reference network, trained for 50 epochs.

\subsubsection{Hardware Platform: GAP8}
A subset of the architectures found by DUCCIO has been deployed on the GreenWaves Technologies' GAP8 IoT multi-core System-on-Chip (SoC), in order to extract real HW metrics such as memory footprint and latency.
The two main functional units of GAP8 are a single 32bits RISC-V core, known as the fabric controller, and a computational cluster made of 8 RISC-V cores.
The fabric controller acts as a microcontroller (MCU) and controls the cluster and other peripherals.
The octa-core cluster runs computationally-intensive tasks, such as DNNs' inference, in a vectorized and parallelized fashion, also thanks to customized instruction set extensions~\cite{dory}.
GAP8 features an internal two-level memory hierarchy.
All cores can reach the 512 kB L2 memory, whereas the
L1 memory is divided into two sections: a 64 kB shared memory for the cluster and 16 kB dedicated to the fabric controller.
If needed a third larger but slower L3 off-chip memory can be reached via a HyperBus interface or a quad-SPI.
To deal with such a complex and heterogeneous memory hierarchy, 
GAP8 additionally features two Direct Memory Access (DMA) co-processors, that can help hiding L2 and L3 memory latency by efficiently moving data within the hierarchy with a double-buffering strategy, while the RISC-V cores perform computations on previously transferred data.
To deploy our networks on GAP8, we converted ONNX graphs exported from PyTorch into C code using the DORY~\cite{dory} compilation toolchain.
Since GAP8 lacks an FPU, the deployed models are quantized to 8-bit, which causes negligible accuracy drops. All results are reported on test sets.

\subsection{Global Constraints}\label{sec:res_global}
Fig.~\ref{fig:pareto_fronts} reports the results obtained applying DUCCIO to both DNAS methods on the MLPerf Tiny tasks.
Each plot reports the reference DNN (black square) and the found architectures (coloured dots) in the Accuracy versus OPs space, using the global constraints formulation of Eq.~\ref{eq:duccio_global}.
Each colour denotes a different storage space constraint $T_{s}$.
We benchmarked our approach by setting $T_{s}$ to be respectively 25\%, 50\% and 75\% of the original size of each reference architecture (we included also 6.25\% and 12.5\% for VWW and AMD, since their reference models are highly over-parametrized, as observed in our previous work~\cite{multiloss_conf}). This setup simulates a real-world deployment scenario with three different MCUs with progressively less available memory.
Within the curve relative to each $T_{s}$ constraint, the right-most point is obtained without any OPs constraint $T_{o}$. The other points are obtained by setting $T_{o}$ to 25\%, 50\% and 75\% of the OPs of that network. Importantly, each colored point in Fig.~\ref{fig:pareto_fronts} is generated, starting from the seed/supernet, with a single DNAS training, given the desired $(T_s, T_o)$ combination.

The leftmost graphs show the results obtained on ICL.
For all the memory and OPs targets, the mask-based DNAS is able to perfectly match the constraints by spanning almost one order of magnitude in OPs, 1.07M-10.0M, and achieving an accuracy between 71.72\% and 85.74\%.
The highest reduction in terms of OPs with a network that matches the accuracy of the baseline (84.31\% vs 84.03\%) is obtained with the path-based approach under the 50\% size constraint and 75\% OPs constraint. This model has 33.3k parameters and 5.68M OPs, 55.9\% and 54.6\% lower than the baseline, respectively.  The path-based DNAS also achieves the best accuracy of 85.7\% while still reducing the number of OPs compared to the seed network by 20\%.

The middle-left graphs report the Pareto fronts for VWW.
For the mask-based NAS, DUCCIO finds networks with similar accuracies and latencies for all constraints, demonstrating that even 6.25\% of the reference size is enough to maximize the network accuracy.
In particular, the found architectures have OPs spanning from 602k to 3.42M and accuracies in the interval 76.1\% - 85.83\%.
Noteworthy, many of them Pareto-dominate the seed: compared to the seed, we achieved the highest accuracy gain of 2.07\% (85.83\% vs 83.76\%) while reducing operations by 54.2\% and memory by 87.4\%.
With the path-based DNAS, in this case, our results span a thinner space, because the reference DNN is mainly composed of depthwise-separable convolutions. Therefore, given the 4 layer alternatives included in our supernets, we can further reduce the memory occupation only by substituting some layers with identities (eliminating them completely). Comparing the graphs, it can be seen that removing layers is not beneficial for this particular task, since the networks found are all Pareto-dominated by the ones found with the mask-based approach. We do not report lower size constraints since they are not achievable within the search space of this second DNAS (the identity layer cannot be selected when the number of channels changes or the stride is $>$ 1).

The middle-right plots show the results on KWS.
With the mask-based DNAS, DUCCIO finds many Pareto optimal solutions for each size constraint.
Notably, with the 75\% size target and without $T_o$, we reach the best accuracy of 92.82\%, only 0.43\% lower compared to the seed, reducing the OPs by 61.37\%.
On the other hand, the path-based DNAS can only find 3 different DNNs, obtained by substituting one, two or three of the four depthwise separable layers of the reference network with identities. Once again, the networks found by the path-based DNAS are outperformed by the mask-based outputs, demonstrating that a finer-grain search is more useful for this particular task.

Lastly, on the rightmost part of Fig.~\ref{fig:pareto_fronts}, we report the results on AMD. As discussed before, due to the nature of the reference network, we present only results with the mask-based DNAS.
Moreover, since the reference is made of FC layers, model size and OPs are directly proportional, making two separate constraints  ($T_s$ and $T_o$) redundant.
For this reason, the plot is composed only of single points and not curves.
Nonetheless, also in this case, DUCCIO is able to find Pareto optimal solutions that fit different deployment scenarios, with limited accuracy drops.

The DUCCIO formulation is orthogonal to the selected DNAS method, hence being applicable to slower/faster algorithms with equal effectiveness. However, since one of the main objectives of our work is to reduce search times, making them effectively comparable to a single training, it is also relevant to report the time costs associated with obtaining the results of Fig.~\ref{fig:pareto_fronts}. As a representative example, we measured the average time overhead for a single training step on the ICL benchmark, with respect to a vanilla training of the reference ResNet architecture. Our training setup is composed of a single NVIDIA GeForce GTX 1080 Ti GPU, and a full ICL reference training requires $\approx$ 2 hours to complete. We obtained that the path-based DNAS has a 2.59x time overhead on average, whereas the mask-based DNAS increases the iteration time by 1.97x. Both values are expected, and motivated respectively by larger supernet size, and by the additional masking operations. Together, they show that the complexity of DUCCIO is easily manageable, even with limited training resources.

\subsection{Layer-wise Constraints}
This section describes the application of DUCCIO to find a network tailored to the memory hierarchy of our HW target, the GAP8 platform.
For this use case, we focus on Tiny ImageNet, since the ResNet18 reference architecture is large enough to show the benefits of layer-wise memory constraints. The target NAS is the mask-based one, which allows for a finer-grain control of the layer sizes.

\begin{figure}[ht]
  \centering
  \includegraphics[width=0.8\columnwidth]{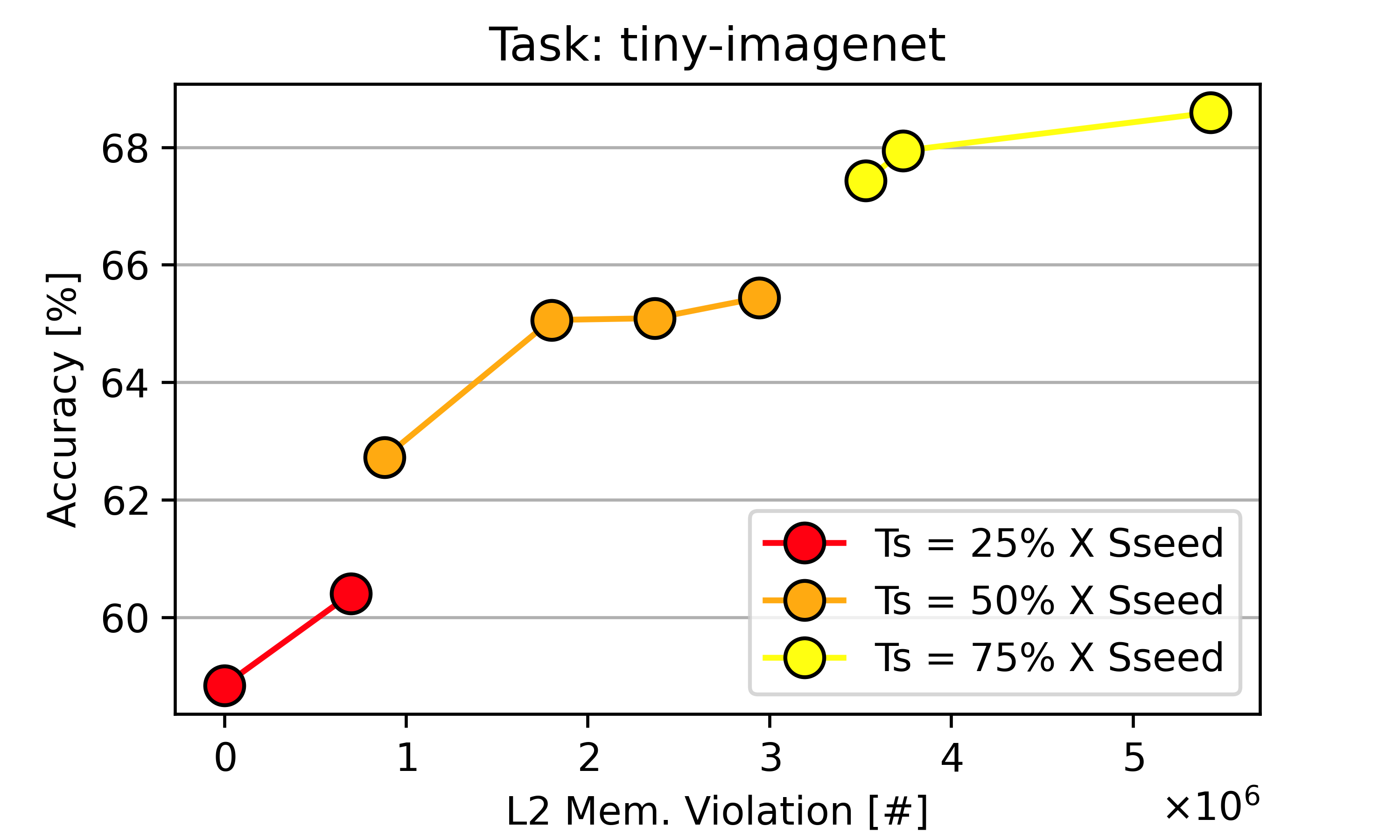}
  \vspace{-0.4cm}
  \caption{Accuracy vs L2 memory violation, i.e., the difference between layer effective memory and L2 dimension.}
  \label{fig:tiny_results}
\end{figure}

The seed has 11.28M parameters and obtains a top-1 accuracy of 71.97\%.
First, we search for three smaller models with $T_s$ equal to 25\%, 50\%, and 75\% of the seed (rightmost points in the red, orange, and yellow curves in Fig. \ref{fig:tiny_results}), corresponding to the availability of a L3 external memory that can store 3M, 6M, and 9M parameters, respectively.
Then, we apply progressively stricter constraints on the layer-wise total memory occupation (Eq.~\ref{eq:duccio_local}). In particular, we consider as critical layers all those layers with $\mathcal{M}^{(n)} < (1+C)T_m$, where $T_m$ is the available L2 memory space, and $C \in \{0.3, 0.6, 0.9\}$.
As a result, we obtain the Pareto frontiers of Fig. \ref{fig:tiny_results}. On the x-axis, we report the total L2 memory violation, computed as $ L2V = \sum_{n \in layers}(\mathcal{M}^{(n)}(\theta) - T_m)$.

\begin{table*}[ht]
\centering
\caption{Deployment of models from ICL on the GAP8 platform. }
\begin{tabular}{|llllllll|}
\label{tab:icl}
\textbf{Model} & \textbf{MOPs Constraint} & \textbf{Memory [kB]} & \textbf{MOPs} & \textbf{Latency [ms]} & \textbf{GOPs/s} & \textbf{Accuracy} & \textbf{Drop w.r.t Ref} \\  \hline\hline
Reference   & None                      & 75.5                 & 12.50           &             21.42          &            0.58     & 84.0 \%  & 0 \%           \\\hline\hline
Mask-Based-L   & None                      & 43.4                 & 9.18           &             21.26          &            0.43     & 85.0 \%    & +1.0 \%         \\
Mask-Based-L   & 6.88                      & 35.6                 & 6.42           &                15.10       &        0.43         & 83.7 \% & -0.3 \%            \\ \hline
Mask-Based-M   & 6.05                      & 32.5                 & 5.68           &                     15.51  &              0.37   & 84.3 \%     & +0.3 \%      \\
Mask-Based-M   & 4.03                      & 24.7                 & 3.80           &     10.68                  &          0.36       & 82.8 \%  & -1.2 \%         \\ \hline
Mask-Based-S   & None                      & 18.7                 & 5.21           &          11.66             &      0.45           & 81.4 \%     & -2.6 \%      \\
Mask-Based-S   & 2.61                      & 17.6                 & 2.60           &               9.40        &   0.28              & 79.0 \%  & -5.0 \%         \\ \hline \hline
Path-Based-L   & 10.0                      & 54.8                 & 10.0          &       17.50               &     0.57            & 85.7 \%     & +1.7 \%      \\
Path-Based-L   & 6.68                      & 51.8                 & 7.55           &        21.76               &            0.35     & 84.1 \%   & +0.1 \%        \\ \hline
Path-Based-M   & 8.19                      & 28.2                 & 7.53           &        17.56               &        0.43         & 83.8 \%      & -0.2 \%     \\
Path-Based-M   & 5.46                      & 16.4                 & 4.42           &           14.11            &      0.31           & 83.7 \%     & -0.3 \%      \\ \hline
Path-Based-S   & None                      & 16.4                 & 4.42           &            14.11           &      0.31           & 83.3 \%      & -0.7 \%     \\
Path-Based-S   & 2.14                      & 14.6                 & 2.48           &            12.59           &   0.20              & 83.1 \%    & -0.9 \%       \\ \hline
\end{tabular}
\end{table*}
\begin{table*}[ht]
\centering
\caption{Deployment of models from Tiny ImageNet on the GAP8 platform after layer-wise memory reduction.}
\begin{tabular}{|llllllll|}
\label{tab:tiny-imagenet}
\textbf{Model} & \textbf{L2V Constraint} & \textbf{Mem. [kB]} & \textbf{MOPs} & \textbf{MCycles} & \textbf{GOPs/s} & \textbf{L2 Violation [kB]} & \textbf{Accuracy} \\ \hline \hline
TinyI-L        & None                    & 7770                 & 1956,8         & 1002.6           & 0.19            & 5301                       & 68.59 \%          \\
TinyI-L-60     & 60 \% $\times$ L2       & 6765                 & 1361,2         & 644.4 (-36\%)    & 0.21 (+8\%)     & 3648                       & 67.94 \%          \\
TinyI-L-90     & 90 \% $\times$ L2       & 6615                 & 1344,3         & 601.1 (-40\%)    & 0.21 (+15\%)    & 3446                       & 67.43 \%          \\ \hline \hline
TinyI-M        & None                    & 5342                 & 1607,6         & 637.3            & 0.25            & 2875                       & 65.44 \%          \\
TinyI-M-30     & 30 \% $\times$ L2       & 4932                 & 1368,1         & 293.9 (-54\%)    & 0.47 (+84\%)    & 2313                       & 65.09 \%          \\
TinyI-M-60     & 60 \% $\times$ L2       & 4706                 & 1164,8         & 291.5 (-55\%)    & 0.40 (+58\%)    & 1758                       & 65.06 \%          \\
TinyI-M-90     & 90 \% $\times$ L2       & 3742                 & 1136,0         & 252.1 (-61\%)    & 0.45 (+79\%)    & 861                        & 62.72 \%          \\ \hline\hline
TinyI-S        & None                    & 2738                 & 1022,3         & 334.5            & 0.31            & 681                        & 60.40 \%          \\
TinyI-S-60     & 60 \% $\times$ L2       & 2264                 & 731,0          & 178.8 (-46\%)    & 0.41 (+34\%)    & 0                          & 58.84 \%          \\ \hline
\end{tabular}
\end{table*}

From this search, we can draw two important conclusions: first, increasing $C$ leads to including more layers in the critical set and, therefore, to reduce L2V. Second, for the same size constraint (for instance, the orange curve), we can reduce the L2V without sacrificing much accuracy.
For instance, with the constraint $T_s$ of 50\%, we indeed observe a reduction of only 0.38\% accuracy for an L2V reduction of 38.9\%, between the rightmost network and the one obtained by setting $C = 0.3$.
In the next section, we will show that the networks with a lower L2V also achieve, in most cases, lower latency and higher efficiency. Noteworthy, this correlation does not always hold since the efficiency also depends on i) the number of layers that violate the constraint and ii) to the nature of the layers, which can influence the efficiency (for example, depthwise layers have lower efficiency on GAP8 with respect to normal convolutions \cite{dory}).

\subsection{Embedded Deployment}
Table~\ref{tab:icl} and Table~\ref{tab:tiny-imagenet} summarize deployment results on the GAP8 SoC for the ICL and Tiny ImageNet benchmarks.
For sake of space, we focus only on these two benchmarks.

Table~\ref{tab:icl} reports ICL models obtained by DUCCIO with the two DNAS methods. For each method, we report two DNNs per size target (75\%, indicated by a -H suffix, 50\%, indicated by -M, and 25\%, indicated by -L).
Namely, we deploy the two rightmost points of each Pareto front of Fig.~\ref{fig:pareto_fronts} which correspond to the Mega OPs (MOPs) constraints detailed in the table. Additionally, for comparison, we also deploy the reference DNN.
The Mem. column reports the memory occupation of each model.
The first thing to notice is that all DNNs satisfy the imposed memory target and MOPs constraints.
Regarding latency, we find solutions that span from 9.40ms to 21.26ms and 12.59ms to 27.86ms, respectively, for the mask-based DNAS and the path-based DNAS.
For DNNs obtained under the same size constraint, we notice that a higher number of OPs implies also a higher latency, showing that the OPs are a sufficiently good latency proxy for our target, when comparing similar networks with the same topology and type of layers.
The only exception is represented by the first Path-Based-L model, which has a higher number of MOPs than the second but is faster (-4.26ms). This is due to the fact that the second solution includes a higher number of depthwise layers with lower arithmetic intensity, which is not taken into account by the OPs metric.

Compared to the reference, we find two interesting ``extreme'' solutions.
With the path-based DNAS, the first -L DNN improves accuracy by 1.7\% while still reducing latency by 1.22$\times$ and memory by 27\%.
Conversely, the second Mask-Based-S network trades an accuracy drop of 5\% with a latency reduction of 2.28$\times$ and a memory footprint reduction of 76.7\%.
These two solutions show that DUCCIO is able to either maximize the accuracy or slightly sacrifice it to fit tighter constraints, depending on the targets provided by the designer.
Furthermore, we observe that almost all the deployed architectures (with the exception of the one described above) incur in an accuracy drop lower than 2.6\% while reducing the latency of up to 2.01$\times$.

Table~\ref{tab:tiny-imagenet} reports the deployment results of the networks obtained on Tiny ImageNet using the DUCCIO's layer-wise size refinement strategy.
We deployed all the DNNs from Fig. \ref{fig:tiny_results}.
First, we observe that for each network, applying our layer-wise size refinement strategy improves the network efficiency in terms of GOPs/s, from +8\% to +84\%, by reducing expensive accesses to L3.
We also observe that, as expected, the L2V metric is correlated to the efficiency, even if not perfectly, for the aforementioned reasons.
Lastly, a tighter constraint (considering more layers as critical) always leads to an improvement in terms of latency, as can be observed in the MCycles column, with reductions from 36\% to 61\%.
Noteworthy, TinyI-M-30 requires 54\% fewer cycles than TinyI-M, improves the GMAC/s efficiency by +84\%, while incurring a negligible accuracy drop (-0.35\%).
\subsection{Ablation Study}
\subsubsection{Constraint vs Objective}
\begin{figure}[ht]
  \centering
  \includegraphics[width=0.8\columnwidth]{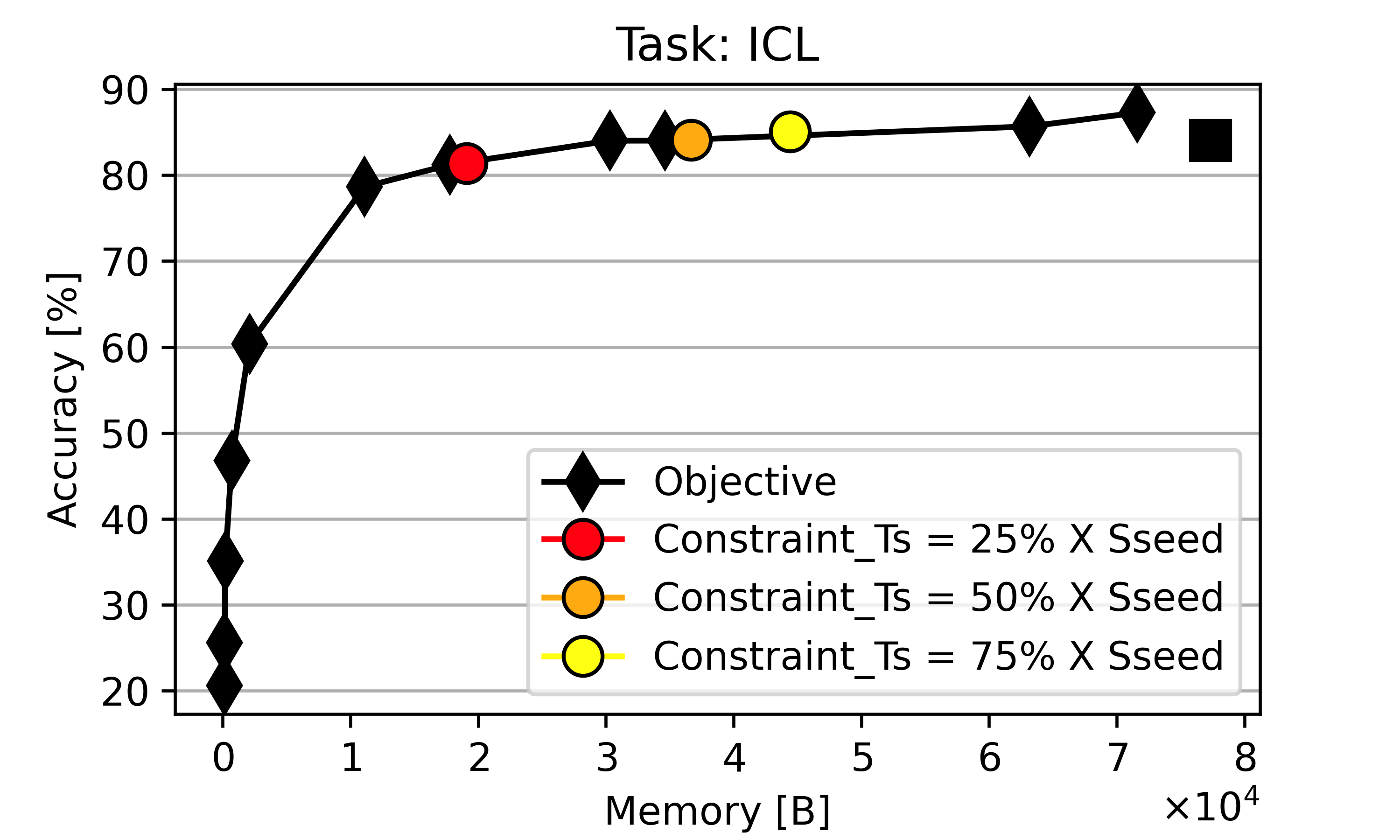}
  \caption{Model size as objective (classical) vs constraint (DUCCIO).}
  \label{fig:constr_obj}
\end{figure}

In Fig. \ref{fig:constr_obj}, we show the main advantage of DUCCIO, that is, the possibility to find a DNN with desired characteristics in one-shot.
For this experiment, we run the mask-based NAS on ICL, both with DUCCIO's formulation and with the classical DNAS formulation of Eq.~\ref{eq:dnas}. In both cases, we consider only the model size as cost metric. For DUCCIO, we set the same size targets as Sec.~\ref{sec:res_global} and no OPs target, hence the results correspond to the rightmost points of the corresponding curves in Fig.~\ref{fig:pareto_fronts}.

With the classical formulation, using twelve different $\lambda$, we find a Pareto curve which spans almost one order of magnitude in memory, with accuracies ranging from 20.64\% up to 87.25\%.
This shows once again that finding a point satisfying a particular memory constraint is not trivial and requires many iterations of the DNAS search.
On the other hand, using DUCCIO, each of the three points is found in a single search, with a different memory target $T_s$.
The networks found by DUCCIO not only respect the constraint but are also Pareto optimal since they all sit on the curve found swiping the classic DNAS's $\lambda$. This shows that our formulation does not degrade the quality of results compared to a classic DNAS.

\subsubsection{Max vs Absolute Value}

\begin{figure}[h]
  \centering
  \includegraphics[width=0.8\columnwidth]{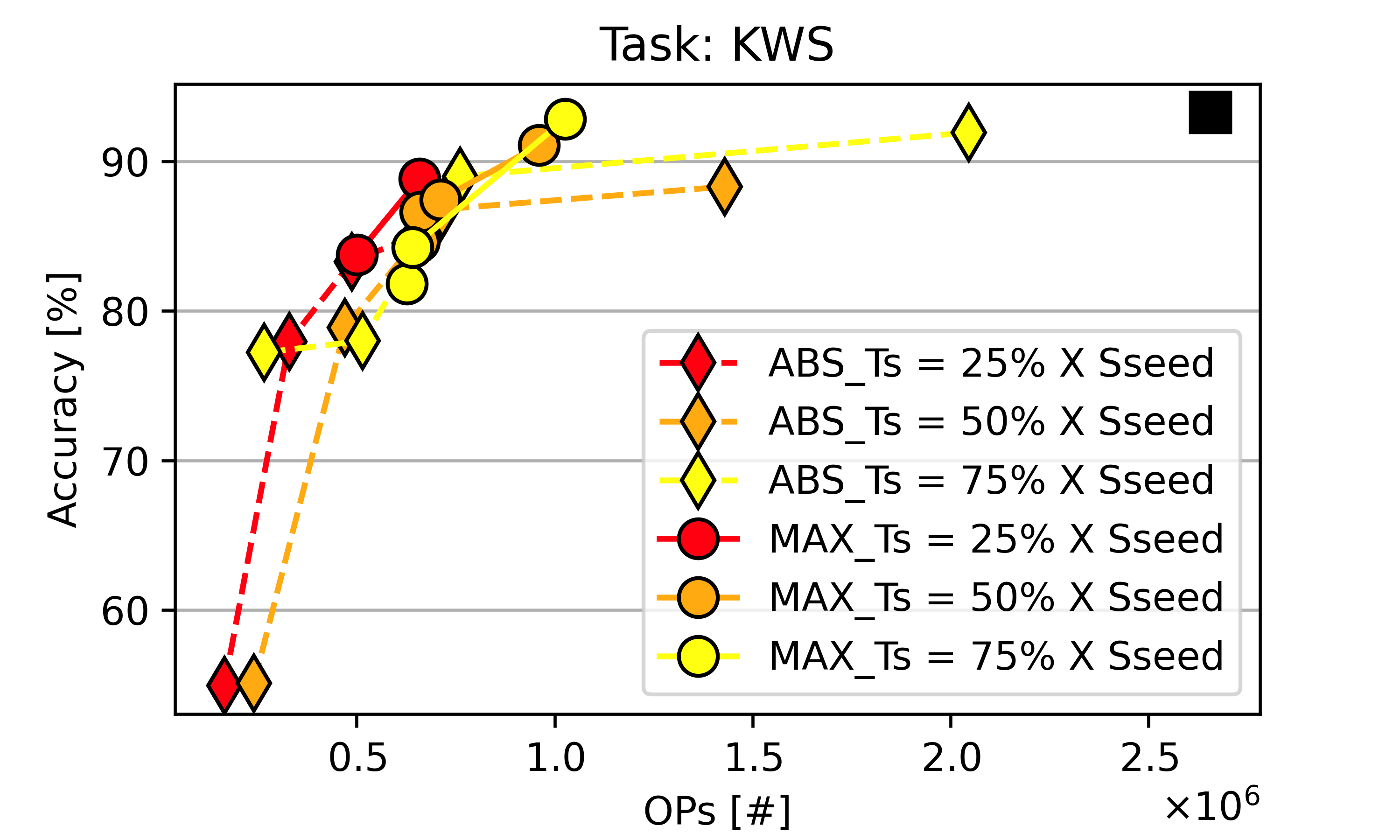}
  \vspace{-0.4cm}
  \caption{Comparison of max() and absolute value constraints for different size targets.}
  \label{fig:abs_max}
\end{figure}
Fig. \ref{fig:abs_max} shows the benefits of using the $\max$ function instead of the absolute value difference from the target in the loss formulation,  as discussed in Sec.~\ref{sec:multireg_loss}.
This experiment is performed on KWS with the mask-based NAS, since the reference network includes an excessive number of channels, making the difference between the two formulations more evident. The circles in the figure correspond to the results of Fig.~\ref{fig:pareto_fronts}, while the diamonds are obtained replacing the max with absolute values in Eq.~\ref{eq:duccio_global}.

Overall, we observe that architectures obtained with the $\max$ function Pareto-dominate the others for relevant accuracies ($>80\%$).
The phenomenon is particularly evident in the two most complex architectures for $T_s =$ 75\% and 50\%.
With the $\max$ function, DUCCIO finds two DNNs with an accuracy of 91.06\% and 92.82\%, and 9.4k and 9.96k parameters, respectively, whereas for the absolute value formulation, under the same $T_s$, the two models achieve 88.30\% and 91.92\% accuracy, with 12.1k and 16.7k parameters. 

In this case, it is evident that forcing the DNAS to produce a model that closely matches the target (using the absolute value formulation) results in two DNNs that are still too over-parametrized for the task, and generalize worse compared to the two counterparts obtained with the $\max$ function.

\subsubsection{Increasing vs Constant $\lambda$}
\begin{figure}[h]
  \centering
  \includegraphics[width=0.8\columnwidth]{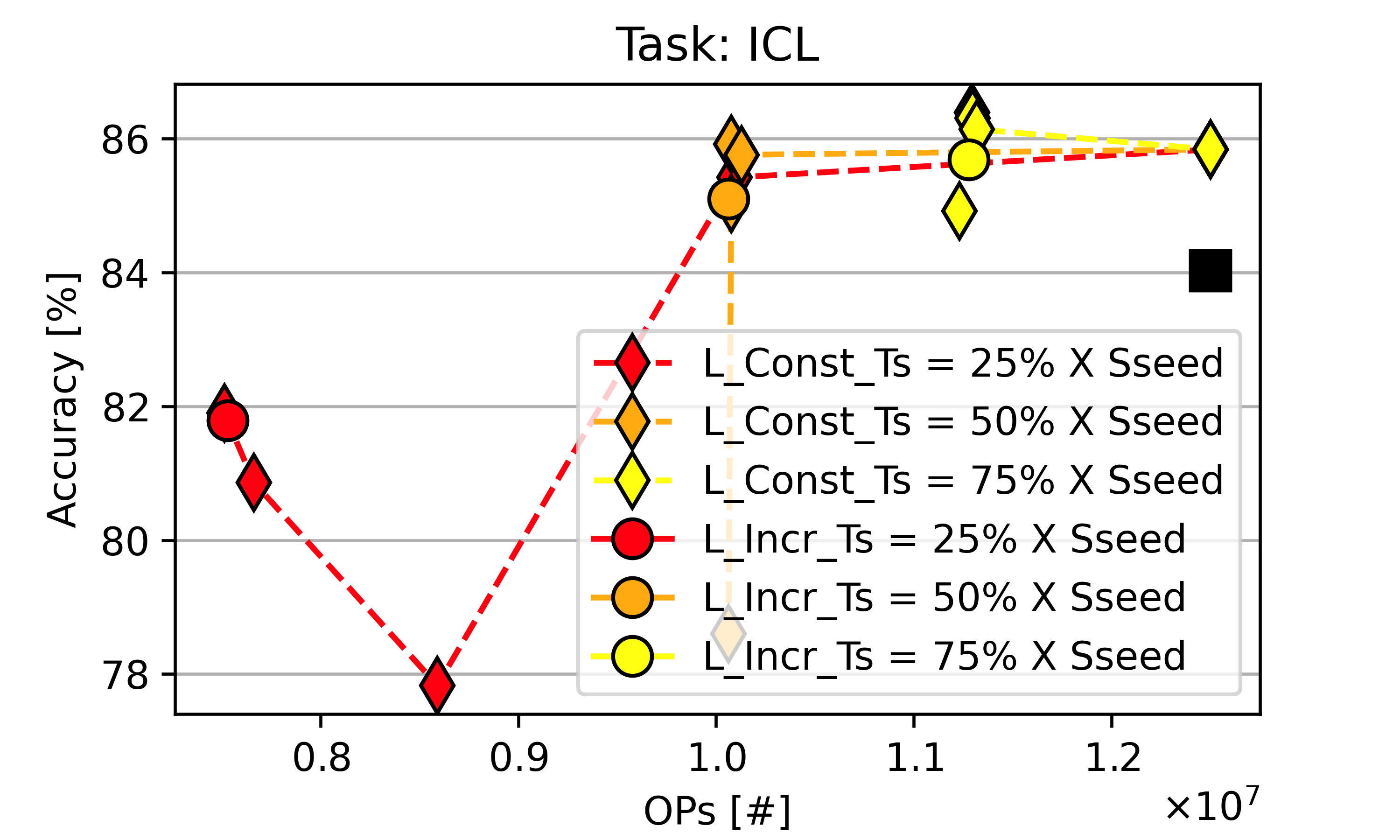}
  \vspace{-0.4cm}
  \caption{Comparison between the scheduling of $\lambda$ proposed in Sec. \ref{sec:training_proc} and a constant $\lambda$ for DUCCIO's training.}
  \label{fig:increasing_const}
\end{figure}

In Fig. \ref{fig:increasing_const}, we show that the $\lambda$ scheduling proposed in Sec. \ref{sec:training_proc} is superior to using a constant regularization strength.
To this end, we perform a search on ICL with the mask-based NAS using DUCCIO with six different constant strengths, both lower and higher than the $\lambda_{target}$ of Eq.~\ref{eq:lambda_target}.
We set once again $T_s$ to 25\%, 50\%, and 75\% of the seed size, and no OPs constraint.

Two critical behaviours can be observed. 
First, when the constant $\lambda$ is too small, the constraint is not respected, leading to networks whose sizes are not reduced at all (e.g., yellow and orange rightmost points). Second, with too large $\lambda$, the shrinking of the networks is too fast, leading to the elimination of channels which are not the least important ones for accuracy. This behaviour is clearly evident on the set of architectures generated with $T_s =$ 50\%, where the output obtained with the largest $\lambda_{const}$ has a loss in accuracy of 6.5\% compared to the one obtained with the scheduling.
Noteworthy, for all $T_s$, the networks found with the scheduling of $\lambda$ reach comparable or superior accuracy compared to solutions found with different constant $\lambda$, while always respecting the imposed constraint.

\subsubsection{Discretized Sampling vs $\theta$ Polarization}
\begin{figure}[ht]
  \centering
  \includegraphics[width=0.8\columnwidth]{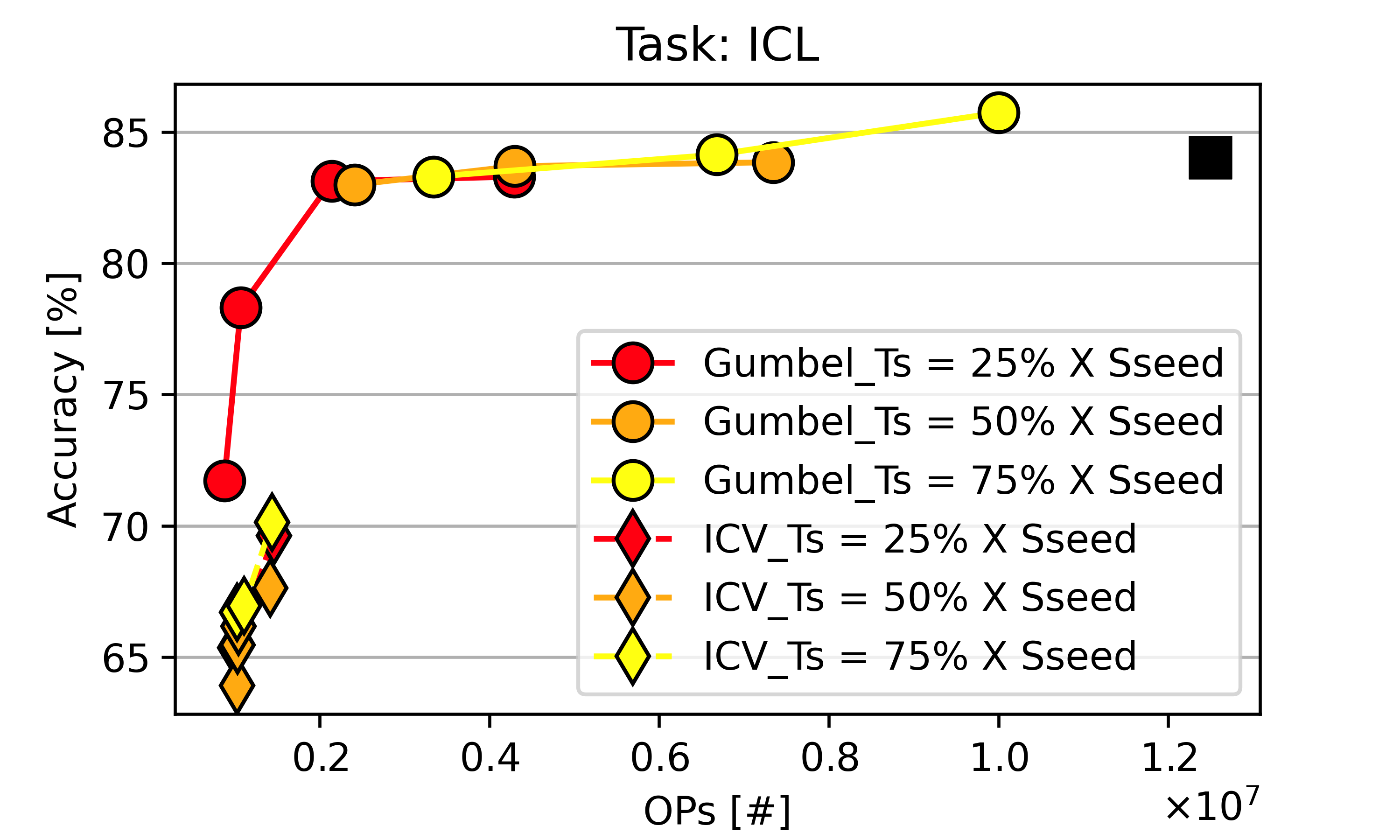}
  \vspace{-0.4cm}
  \caption{Comparison between using ICV Loss to polarize $\theta$ coefficients and discretized Gumbel Softmax sampling.}
  \label{fig:icv_gumbel}
\end{figure}

In Fig. \ref{fig:icv_gumbel}, we motivate the last choice of our training recipe, that is, using the Gumbel Softmax trick with discretized sampling to choose one of the alternative layers in the path-based DNAS, rather than a soft sampling with $\theta$ coefficients polarized through an ICV loss. The experiment is run again on ICL, with the same targets of Fig.~\ref{fig:pareto_fronts}.

The difference between the two approaches is evident. The main downside of using the ICV loss is that all alternative layers' outputs are still contributing to the global output of a supernet module, although each with different magnitude, due to the polarized coefficients. This leads to weights co-adaptation~\cite{1000nas}, which means that the weights adapt to the fact that the output of the module depends on all alternatives simultaneously, leading to a very poor accuracy when the architecture is discretized at the end of search, despite the following finetuning.

\section{Conclusion}\label{sec:conclusion}
In this work, we proposed DUCCIO, a new DNAS formulation that allows employing multiple hardware-related constraints to find the desired network in one shot.
We applied DUCCIO to two different DNAS methods, one mask-based and one path-based, testing on five benchmarks.
We found networks that reach the baseline accuracy while reducing the number of parameters and OPs by 55.9\% and 54.6\% on ICL, or that outperform the baseline model by 2.07\% accuracy on VWW, while reducing OPs by 54.2\% and memory by 87.4\%.
Deploying ICL models, we have shown that our results are well correlated to the final performance, showing an improvement in accuracy of 1.7\%, while reducing latency by 1.22$\times$ and memory by 27\%.

\bibliographystyle{IEEEtran}
\bibliography{IEEEabrv,bibliography}


\begin{IEEEbiography}[{\includegraphics[width=1in,height=1.25in,clip,keepaspectratio]{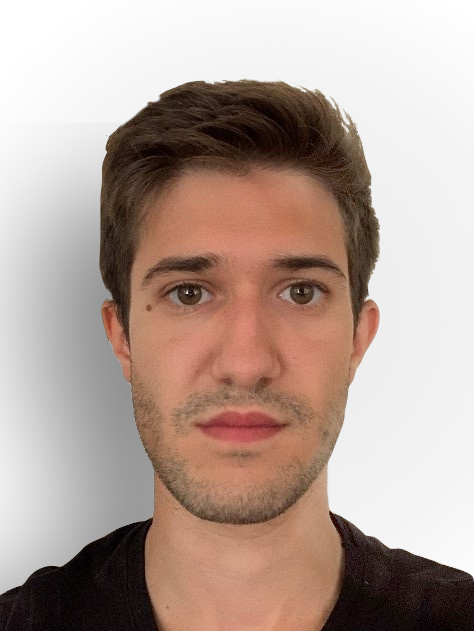}}]{Alessio Burrello}
received his M.Sc. and Ph.D degrees in Electronic Engineering at the Politecnico of Turin, Italy, and University of Bologna, respectively, in 2018 and 2023.
He is currently a research assistant at Politecnico di Torino.
His research interests include parallel programming models for embedded systems, machine and deep learning, hardware-oriented deep learning, and code optimization for multi-core systems. 
\end{IEEEbiography}

\begin{IEEEbiography}[{\includegraphics[width=1in,height=1.25in,clip,keepaspectratio]{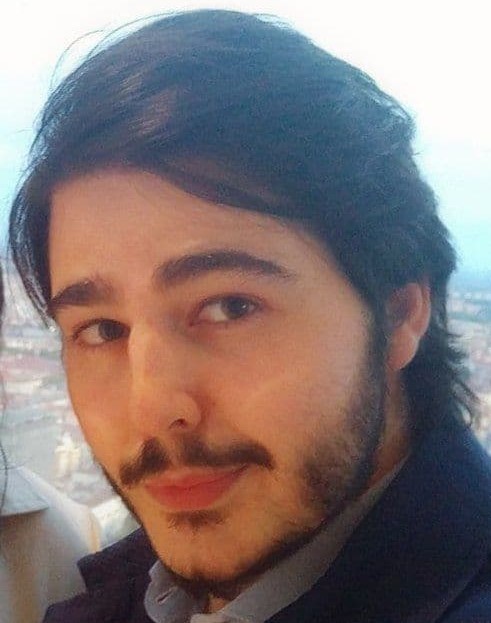}}]{Matteo Risso}
received his B.Sc degree in Physical Engineering and M.Sc degree in Electronic Engineering at the Politecnico di Torino, Italy, in 2018 and 2020. He is currently working toward his Ph.D. degree at Politecnico di Torino, Italy. His research interests include Embedded Machine Learning and Energy-Efficient Embedded Systems.
\end{IEEEbiography}

\begin{IEEEbiography}[{\includegraphics[width=1in,height=1.25in,clip,keepaspectratio]{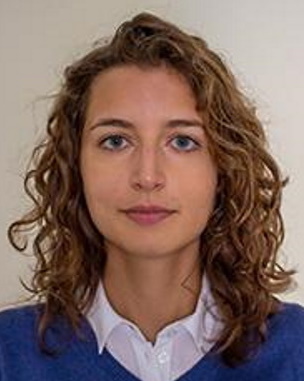}}]{Beatrice Alessandra Motetti} received her B.Sc. degree in Computer Engineering and her M.Sc. degree in Data Science and Engineering at Politecnico di Torino, Italy, in 2020 and 2022 respectively. She is currently a Research Assistant at Politecnico di Torino. Her research interests include Neural Architecture Search and Representation Learning.
\end{IEEEbiography}

\begin{IEEEbiography}[{\includegraphics[width=1in,height=1.25in,clip,keepaspectratio]{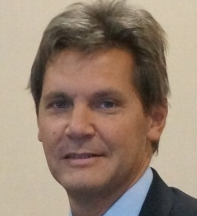}}]{Enrico Macii} is a Full Professor of Computer Engineering with the Politecnico di Torino, Torino, Italy. He holds a Laurea degree in electrical engineering from the Politecnico di Torino, a Laurea degree in computer science from the Universita'  di Torino, Turin, and a PhD degree in computer engineering from the Politecnico di Torino. His research interests are in the design of digital electronic circuits and systems, with a particular emphasis on low-power consumption aspects energy efficiency, sustainable urban mobility, clean and intelligent manufacturing. He is a Fellow of the IEEE.
\end{IEEEbiography}

\begin{IEEEbiography}[{\includegraphics[width=1in,height=1.25in,clip,keepaspectratio]{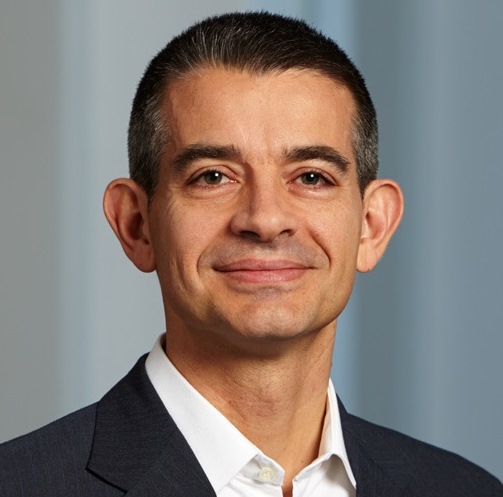}}]{Luca Benini}
is the Chair of Digital Circuits and Systems at ETH Z\"urich and a Full Professor at the University of Bologna.
He has served as Chief Architect for the Platform2012 in STMicroelectronics, Grenoble.
Dr. Benini’s research interests are in energy-efficient systems and multi-core SoC design. 
He is also active in the area of energy-efficient smart sensors and sensor networks. 
He has published more than 1’000 papers in peer-reviewed international journals and conferences, five books and several book chapters. 
He is a Fellow of the ACM and a member of the Academia Europaea.
\end{IEEEbiography}

\begin{IEEEbiography}[{\includegraphics[width=1in,height=1.25in,clip,keepaspectratio]{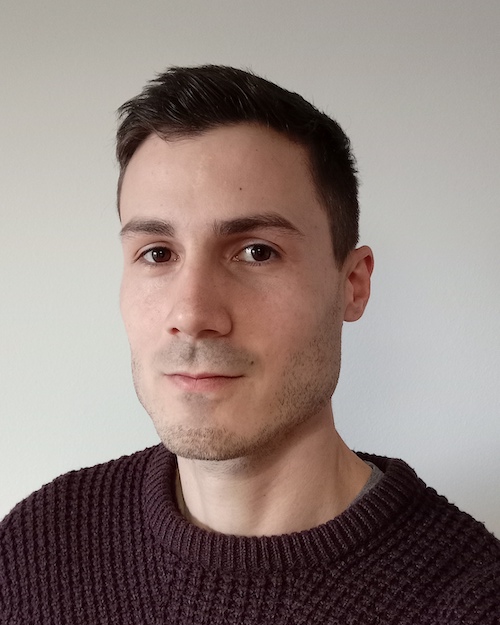}}]{Daniele Jahier Pagliari} received the M.Sc. and Ph.D. degrees in computer engineering from the Politecnico di Torino, Turin, Italy, in 2014 and 2018, respectively. He is currently an Assistant Professor with the Politecnico di Torino. His research interests are in the computer-aided design and optimization of digital circuits and systems, with a particular focus on energy-efficiency aspects and on emerging applications, such as machine learning at the edge.
\end{IEEEbiography}

\vfill

\end{document}